\documentclass[runningheads]{llncs}

\usepackage{eccv}

\usepackage{eccvabbrv}

\usepackage{graphicx}
\usepackage{booktabs}

\usepackage[accsupp]{axessibility}  

\usepackage{xcolor}
\usepackage{colortbl}
\definecolor{LightGrey}{rgb}{0.92,0.92,0.92}
\usepackage[misc]{ifsym}

\usepackage{orcidlink}

\begin{document}

\title{Open Vocabulary 3D Scene Understanding via Geometry Guided Self-Distillation} 

\titlerunning{GGSD}

\author{Pengfei Wang\inst{1,2}\orcidlink{0000-0002-3675-9508} \and
Yuxi Wang\inst{2}\orcidlink{0000-0003-1579-2357} \and
Shuai Li\inst{1}\orcidlink{0000-0003-0760-5267}  \and
Zhaoxiang Zhang\inst{1,2,3,4}\textsuperscript{(\Letter)}  \and
Zhen Lei\inst{1,2,3,4}  \and
Lei Zhang\inst{1}\textsuperscript{(\Letter)}\orcidlink{0000-0002-2078-4215}}

\authorrunning{P. Wang et al.}

\institute{The Hong Kong Polytechnic University \and
Centre for Artificial Intelligence and Robotics, HKISI, CAS 
 \and
State Key Laboratory of Multimodal Artificial Intelligence Systems, CASIA \and
School of Artificial Intelligence, University of Chinese Academy of Sciences (UCAS)
\\ 
\email{pengfei.wang@connect.polyu.hk, zhaoxiang.zhang@ia.ac.cn, cslzhang@comp.polyu.edu.hk}}


\maketitle

\begin{abstract}
The scarcity of large-scale 3D-text paired data poses a great challenge on open vocabulary 3D scene understanding, and hence it is popular to leverage internet-scale 2D data and transfer their open vocabulary capabilities to 3D models through knowledge distillation. 
However, the existing distillation-based 3D scene understanding approaches rely on the representation capacity of 2D models, disregarding the exploration of geometric priors and inherent representational advantages offered by 3D data.
In this paper, we propose an effective approach, namely Geometry Guided Self-Distillation (GGSD), to learn superior 3D representations from 2D pre-trained models. 
Specifically, we first design a geometry guided distillation module to distill knowledge from 2D models, and then leverage the 3D geometric priors to alleviate the inherent noise in 2D models and enhance the representation learning process. Due to the advantages of 3D representation, the performance of the distilled 3D student model can significantly surpass that of the 2D teacher model. This motivates us to further leverage the representation advantages of 3D data through self-distillation.
As a result, our proposed GGSD approach outperforms the existing open vocabulary 3D scene understanding methods by a large margin, as demonstrated by our experiments on both indoor and outdoor benchmark datasets. Codes are available at \url{https://github.com/Wang-pengfei/GGSD}.

\end{abstract}

\section{Introduction}
\label{sec:intro}
3D scene understanding plays a vital role in various real-world applications, including robot manipulation, virtual reality, and autonomous driving, \etc.
Traditional 3D scene understanding methods mostly employ supervised training on labeled 3D datasets with a limited number of classes~\cite{graham20183d,vu2022softgroup,misra2021-3detr}. 
These approaches, however, face difficulties when they encounter unknown categories in open environments. Such a limitation in practical applications motivates researchers to explore new 3D scene understanding models with open vocabulary capability. 
Inspired by the success of vision-language foundation models like CLIP~\cite{radford2021learning}, which aligns language features to image features, in various image understanding tasks~\cite{gu2021open,Hanoona2022Bridging,xu2021simple,li2022languagedriven,zhou2022maskclip,zhang2022pointclip,huang2022clip2point}, an intuitive strategy is to leverage a substantial amount of paired 3D point cloud and text data for training. Unfortunately, the high cost of annotating 3D datasets (\eg, annotating a scene with 20 categories takes approximately 22.3 minutes~\cite{ding2023pla}) makes it impractical to manually label enough data with numerous real-world categories.

\begin{figure}[t] 
\centering
   \includegraphics[width=0.55\linewidth]{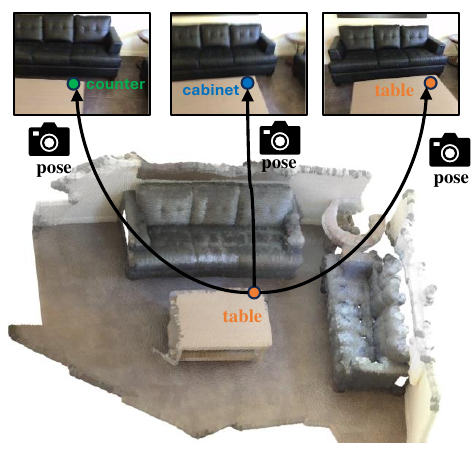}
\caption{Example of the problems when utilizing 2D models, \eg, LSeg~\cite{li2022languagedriven}, for pixel-to-point distillation. We can see the noise caused by the factors such as occlusion, lighting variations and view angles in 2D images.
Therefore, only employing the 2D model in distillation brings errors to the 3D model, limiting its 3D representation capacity.
}
\label{fig:intro}
\end{figure}

An alternative yet more practical approach is to leverage internet-scale 2D data to acquire open vocabulary capability, and then transfer this capability to 3D scene understanding models via distillation~\cite{ding2023pla,peng2023openscene,yang2023regionplc,zhang2023clip,chen2023clip2scene}. 
For instance, 
PLA~\cite{ding2023pla} utilizes a 2D caption model~\cite{wang2022ofa} to generate textual descriptions for images, which are then refined using GPT2~\cite{vit-gpt2}.
Subsequently, these descriptions are projected onto the 3D point cloud, and the vision-language knowledge can be passed to the 3D model via 3D-text contrastive learning.
The recent state-of-the-art OpenScene~\cite{peng2023openscene} utilizes 2D open-vocabulary segmentors such as LSeg~\cite{li2022languagedriven} and OpenSeg~\cite{ghiasi2022scaling} to extract pixel-level 2D embeddings for the 3D point cloud. The ability of pixel-text alignment is then transferred to align 3D point to text via feature distillation, enabling 3D semantic-level scene understanding.

Though the above mentioned distillation-based methods have shown promising results in 3D scene understanding, they basically imitate the 2D models and inherit their limitations, restricting the upper bound of representation power. 
As shown in Fig. \ref{fig:intro}, the 2D model may suffer from errors due to issues such as occlusion, lighting variations, and viewing angles.
The misclassification of extracted 2D features leads to erroneous supervisory signals during the distillation process of 3D point cloud learning. Instead of solely imitating 2D models, our objective is to harness both the geometric priors in 3D point clouds and the representational advantages of 3D data to guide the knowledge distillation process from pre-trained 2D models. 

To achieve the above mentioned objectives, in this paper we propose an effective approach, namely Geometry Guided Self-Distillation (GGSD), to learn superior 3D representations from 2D pre-trained open vocabulary segmentors.
Given a sparse point cloud composed of multiple semantic categories, geometrically homogeneous neighborhoods typically correspond to the same semantic information. Based on this observation, we first design a geometry guided distillation module to learn knowledge from 2D pre-trained models. Specifically, we generate geometrically homogeneous neighborhoods as superpoints and utilize their semantic consistency to guide the distillation process.
The utilization of 3D geometric priors could mitigate the inherent noise of the 2D model and enhance the representation learning capability.
Furthermore, due to the representation advantages of 3D data, which are less affected by factors such as lighting and viewing angles, the distilled 3D student model can exhibit superior performance compared to the 2D teacher model. This motivates us to further distill knowledge within the 3D network itself through self-distillation.
We employ an exponential moving average (EMA) model to predict each 3D point cloud and perform voting within the superpoint to provide supervision signals for the online 3D model. This approach ensures that the online model could receive reliable and accurate guidance by leveraging the consensus among multiple-point predictions.
Finally, our proposed  GGSD method could learn robust 3D representations from the 3D distillation model itself, achieving significantly superior performance to the 2D teacher model.


The contributions of this work are summarized as follows. First, we analyze the limitations of pixel-point distillation methods and propose leveraging the advantages of 3D data representation through self-distillation, instead of relying solely on 2D representations.
Secondly, we utilize 3D geometric priors to mitigate the noise in pixel-point distillation process, thereby enhancing the capability of representation learning.
Third, we conduct experiments on both indoor and outdoor datasets, including ScanNet~\cite{dai2017scannet}, Matterport~\cite{chang2017matterport3d}, and nuScenes~\cite{caesar2020nuscenes}. The results demonstrate that GGSD sets a new state-of-the-art for 3D open vocabulary scene understanding in both indoor and outdoor datasets.


\section{Related Work}
\noindent\textbf{Closed-set 3D Scene Understanding.} 
Previous studies in closed-set 3D scene understanding ~\cite{choy20194d,han2020occuseg,hu2021bidirectional,hu2021vmnet,huang2019texturenet,li2022panoptic,nekrasov2021mix3d,qi2017pointnet++,robert2022learning} have led to significant advancements across diverse 3D scene understanding benchmarks, encompassing tasks such as 3D object classification, 3D object detection, 3D semantic and instance segmentation, \etc.
Some researches~\cite{chen2023pimae} have explored the use of 2D pre-trained models, such as MAE~\cite{he2022masked}, for pre-training point encoders. However, they mainly emphasize on leveraging pre-training to enhance supervised 3D downstream tasks, rather than addressing open-vocabulary query scenarios.
Existing closed-set 3D scene understanding methods excel in specific categories but often struggle to recognize new categories in open-set scenarios. Our objective is to develop a solution to effectively recognize and comprehend new categories in open-set environments, thus overcoming this limitation.

\noindent\textbf{Open-Vocabulary 2D Scene Understanding.}
Traditional closed-set scene understanding has seen significant performance improvements~\cite{cheng2022masked,zhang2023mp,fan2022toward,wang2021uncertainty,wang2023informative,wang2023pulling}, while open-vocabulary scene understanding is also beginning to flourish.
Prior studies have explored knowledge distillation from visual-language models, such as CLIP, to transfer their open vocabulary capabilities in 2D vision tasks ~\cite{Du2022LearningTP,Feng2022PromptDetTO,Ghiasi2021OpenVocabularyIS,Gu2021OpenvocabularyOD,Li2022LanguagedrivenSS,Liang2022OpenVocabularySS,Lddecke2021ImageSU,Rasheed2022BridgingTG,Zang2022OpenVocabularyDW,Zhong2021RegionCLIPRL}. Recent research endeavors have aimed to establish correlations between dense image features and embeddings derived from language models~\cite{ghiasi2022scaling,li2022languagedriven,gu2021open,kuo2022f,liang2023open,rao2022denseclip}. Due to their pixel-level text alignment capabilities, 2D open vocabulary scene understanding models, such as OpenSeg~\cite{ghiasi2022scaling} and LSeg~\cite{li2022languagedriven}, can serve as teacher models for distillation of 3D models compared to CLIP. In this study, we rely on pre-trained 2D models with open vocabulary capabilities, specifically OpenSeg and LSeg, and propose a novel geometry-guided self-distillation strategy to conduct 3D scene-level understanding tasks without the need for any ground-truth 2D or 3D training data. 

\begin{figure*}[!t]
\begin{center}
\includegraphics[width=0.98\linewidth]{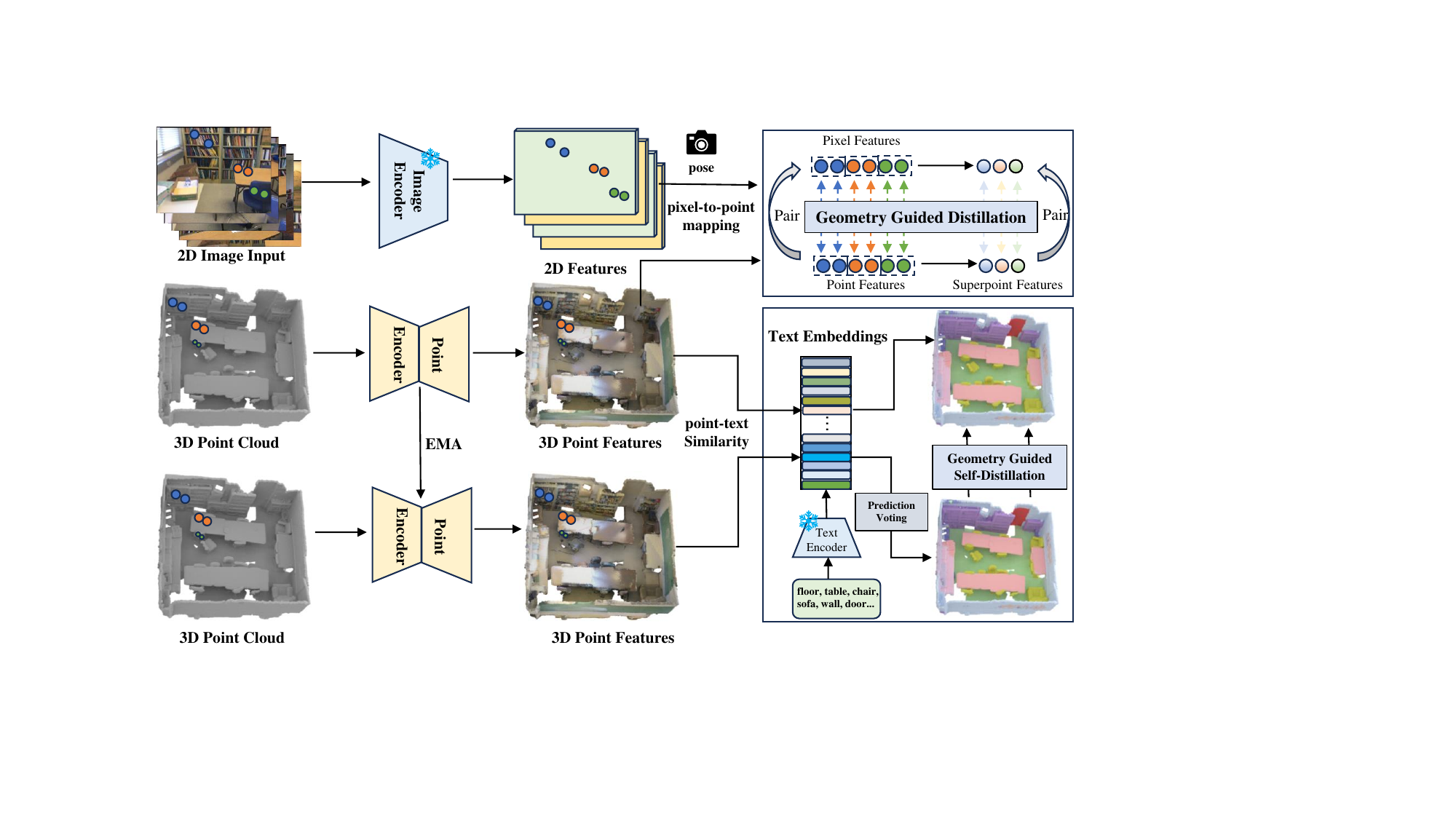}
\end{center}
    \caption{Overview of our Geometry Guided Self-Distillation (GGSD) framework, which consists of two main components: Geometry Guided Distillation and Self-Distillation. The first module leverages 3D geometric priors to mitigate the inherent noise in 2D models.  In the second module, the 3D network learns from its own predictions. This self-distillation step allows the 3D network to continuously enhance its understanding and representation capabilities.
Overall, the GGSD framework combines geometry guidance and self-distillation to fully exploit the valuable information hidden in 3D data, rather than solely relying on 2D pre-trained models.} 
    \label{fig:framework}
\end{figure*}

\noindent\textbf{Open-Vocabulary 3D Scene Understanding.}
Recent works on open-vocabulary 3D scene understanding can be classified into two categories. The first category \cite{qin2023langsplat,kerr2023lerf,liu2024weakly, hu2024semantic}, represented by methods like LERF \cite{kerr2023lerf}, focuses on optimization-based approaches that utilize techniques such as NeRF \cite{mildenhall2021nerf} or 3DGS \cite{kerbl20233d} to distill 2D features into 3D scenes. However, these methods lack generalization capability when confronted with new scenes. The second category~\cite{ding2023pla, chen2023clip2scene, peng2023openscene, takmaz2023openmask3d}, represented by methods like OpenScene~\cite{peng2023openscene}, is based on learning-based approaches. This paper falls in the second category.
CLIP2Scene~\cite{chen2023clip2scene} leverages CLIP image model to extract 2D image features and utilizes textual category features to classify pixels, which are then projected onto 3D point clouds. OpenScene~\cite{peng2023openscene} aggregates pixel features from multiple view images on every 3D point, achieving state-of-the-art open vocabulary 3D scene understanding performance.
Despite the promising results of these methods, their simple imitation of 2D models limits the representational capacity of 3D models.


\section{Method}~\label{sec:method}
An overview of our approach is illustrated in Fig. \ref{fig:framework}.   
Firstly, we follow OpenScene~\cite{peng2023openscene} to create pixel-point feature pairs for the 3D point cloud (see Section \ref{sec:fusion}).
Subsequently, we analyze the limitations of pixel-point distillation methods (see Section \ref{sec:rethink}).
Then, we propose a geometry guided distillation module to enhance the learning of 3D representations from the 2D model (see Section \ref{sec:ggd}).
Finally, we introduce a geometry guided self-distillation module to further distill the knowledge within the 3D network itself (see Section \ref{sec:ggsd}).

\subsection{Creating Pixel-Point Feature Pairs}
\label{sec:fusion}

 Following OpenScene~\cite{peng2023openscene}, our approach begins by extracting dense per-pixel embeddings from RGB images using a pre-trained 2D vision language segmentation model, such as LSeg~\cite{li2022languagedriven} or OpenSeg~\cite{ghiasi2022scaling}. 
 After obtaining the pixel features, we can establish correspondences between each point in the point cloud and the pixels in different RGB images by utilizing the intrinsic matrix and the world-to-camera extrinsic matrix.
Next, we employ a multi-view fusion strategy to combine the features from multiple views into a single feature vector, denoted as $\mathbf{f}^{\text{2D}} = \phi(\mathbf{f}_1, \cdots, \mathbf{f}_K)$, for each point cloud $\mathbf{p}$. This fusion is achieved using an average pooling operator $\phi$ over the $K$-view features.
After obtaining the fused 2D features for each of the M points in the scene, we obtain a 2D pixel feature set $\mathbf{F}^{\text{2D}} = {\mathbf{f}_1^{\text{2D}}, \cdots, \mathbf{f}_M^{\text{2D}}}\in\mathbb{R}^{M\times C}$ corresponding to the point cloud scene. Finally, these embedded features are back-projected into the 3D point cloud scene.
For more details, please refer to OpenScene~\cite{peng2023openscene}.

\subsection{Limitations of Pixel-Point 3D Distillation}
\label{sec:rethink}
Due to the text-image alignment capability of CLIP, the 2D features $\mathbf{F}^{\text{2D}}$ can be directly utilized for language-driven open vocabulary scene understanding. 
To transfer 2D vision-language knowledge to a 3D point cloud network, the distillation approach has been popularly used. One up-to-date pixel-point 3D distillation work is OpenScene~\cite{peng2023openscene}.
Specifically, OpenScene aims to learn a 3D encoder that extracts 3D features from each point cloud, which are expected to be similar to their corresponding 2D pixel features. This allows the 3D encoder to distill the open vocabulary capability of the 2D model. To this end, OpenScene utilizes cosine distance to perform pixel-point distillation:
\begin{equation}
    \mathcal{L}_{p} = 1 - \text{cos}(\mathbf{F}^\text{2D}, \mathbf{F}^\text{3D}).
    \label{eq:loss}
\end{equation}

However, the current pixel-to-point distillation methods are mostly a simplistic imitation of the 2D models, which are limited in extracting useful information from 3D point data. 
Firstly, relying solely on the projection relationship between point clouds and pixels, these approaches fail to fully exploit the geometric priors inherent in 3D data, leading to suboptimal performance.
In this paper, we propose to incorporate 3D geometric priors throughout the entire distillation process to address the issue of noise in 2D models by constraining semantic consistency.
Secondly, the current pixel-point distillation methods fail to fully exploit the representational advantage offered by 3D point clouds. 
It has been observed that the distilled 3D model can outperform the directly projected 2D models.
This can be attributed to the inherent capability of 3D data to handle challenges like lighting variations, occlusion, and viewpoint changes, which are prevalent in 2D representations. Therefore, the accuracy of 3D model segmentation results, particularly for foreground classes, can be significantly improved.
This observation provides a strong motivation for us to further leverage the advantages of 3D representation within the 3D network itself via self-distillation, rather than  relying solely on information from the 2D model.

\subsection{Geometry Guided Distillation}\label{sec:ggd}

\begin{figure}[t]
\setlength{\abovecaptionskip}{ 2 pt}
\setlength{\belowcaptionskip}{ -10 pt}
\centering
   \includegraphics[width=0.75\linewidth]{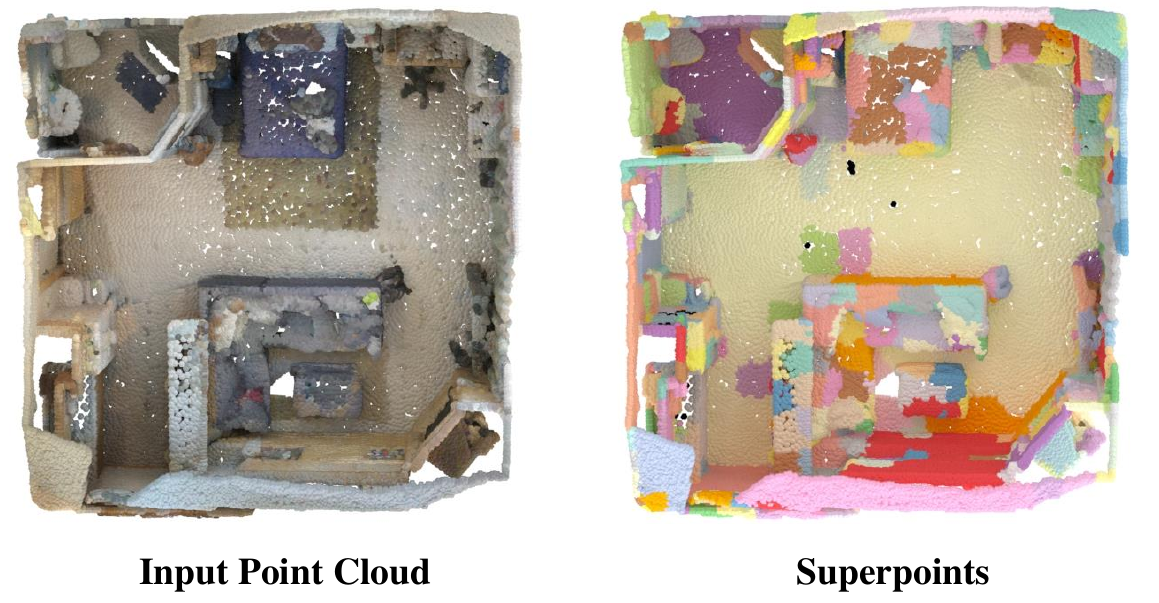}
\caption{Example of the superpoints generated using VCCS \cite{papon2013voxel} on ScanNet scene point clouds. Each colored patch corresponds to a distinct superpoint, showcasing a simple geometric structure. Additionally, the points within each superpoint consistently exhibit a shared semantic category.}
\label{fig:superpoint}
\end{figure}

We first design a geometry guided distillation method to learn knowledge from 2D models.
Given a sparse point cloud composed of multiple semantic categories, we can decompose it into many simple parts, \ie, superpoints, according to their geometric position in the space. As shown in Fig. \ref{fig:superpoint}, the superpoints for an indoor room are constructed by VCCS \cite{papon2013voxel}.  The details for creating superpoints can be found in the \textbf{supplementary file}.
 For a specific input point cloud $\mathbf{P} \in \mathbb{R}^{M\times 3}$, its superpoints are denoted as $\{\mathbf{\Tilde{p}}_1 \cdots \mathbf{\Tilde{p}}_{n} \cdots \mathbf{\Tilde{p}}_{N}\}$, where each superpoint $\mathbf{\Tilde{p}}_{n}$ consists of a small subset of original point cloud $\mathbf{p}$. These superpoints are geometrically simple and usually cannot cover objects of different classes. Ideally, each superpoint as a whole should belong to the same semantic category.
The simplicity and coherence of superpoints naturally facilitate the implementation of semantic consistency constraints.

To mitigate the noise issue in 2D models, we can leverage the semantic consistency provided by 3D geometric priors.
In particular, for a specific input point cloud  $\mathbf{P}$, we have its neural features  $\mathbf{F}^\text{3D} = \{\mathbf{f}_1^{\text{3D}}, \cdots, \mathbf{f}_M^{\text{3D}}\}$, the corresponding pixel fusion  features $\mathbf{F}^{\text{2D}} = \{\mathbf{f}_1^{\text{2D}}, \cdots, \mathbf{f}_M^{\text{2D}}\}\in\mathbb{R}^{M\times C}$ and superpoints $\{\mathbf{\Tilde{p}}_1 \cdots \mathbf{\Tilde{p}}_{n} \cdots \mathbf{\Tilde{p}}_{N}\}$. Firstly, we compute the mean 2D pixel features for superpoints, denoted as $\mathbf{F}^\text{2D}_{sp} = \{\mathbf{\Tilde{f}}_1^{\text{2D}}, \cdots, \mathbf{\Tilde{f}}_N^{\text{2D}}\}$:
\begin{equation}\label{eq:meth_featSP}
    \mathbf{\Tilde{f}}_n^{\text{2D}} = \frac{1}{Q}\sum_{q=1}^Q \mathbf{f}_q^{\text{2D}}, \quad \mathbf{f}_q^{\text{2D}}\in \mathbb{R}^{1\times C},
\end{equation}
where $Q$ is the total number of 3D points within a superpoint $\mathbf{\Tilde{p}}_{n}$, and $\mathbf{f}_q^{\text{2D}}$ is the feature vector for the $q^{th}$ pixel that corresponds to the 3D point in the superpoint. 
Similarly, we can employ the same approach to calculate the mean 3D point features  $\mathbf{F}^\text{3D}_{sp} = \{\mathbf{\Tilde{f}}_1^{\text{3D}}, \cdots, \mathbf{\Tilde{f}}_N^{\text{3D}}\}$. 

To enforce the superpoint feature $\mathbf{F}^\text{3D}_{sp}$ to be consistent with the 2D fused features $\mathbf{F}^\text{2D}_{sp}$, we use a cosine similarity loss as that in pixel-point distillation:
\begin{equation}
    \mathcal{L}_{sp} = 1 - \text{cos}(\mathbf{F}^\text{2D}_{sp}, \mathbf{F}^\text{3D}_{sp}).
    \label{eq:loss_sp}
\end{equation}
The final optimization objective for geometry guided distillation is as follows:
\begin{equation}
    \mathcal{L}_{d} =  \mathcal{L}_{p} +  \mathcal{L}_{sp}.
    \label{eq:loss_gsd}
\end{equation}

\subsection{Geometry Guided Self-Distillation}
\label{sec:ggsd}

With geometry guided distillation, the performance of the 3D model can surpass that of the directly projected 2D model. 
This inspires us to further design a self-distillation module to fully leverage the inherent advantages of 3D data in representation.
Recent methods such as CNS~\cite{chen2023towards} and CLIP2Scene~\cite{chen2023clip2scene} typically transfer the pixel-level pseudo-labels from the 2D image to the 3D point cloud using a 2D-3D calibration matrix. 
In contrast to existing methods that rely on projecting from 2D, we introduce a geometry guided self-distillation approach to distill knowledge from the 3D model itself. Specifically, we generate predictions directly on 3D point clouds instead of projecting 2D predictions back to 3D, as our 3D model possesses enhanced feature representation capabilities. Specifically, given the 3D point features $\mathbf{F}^{\text{3D}} = \{\mathbf{f}_1^{\text{3D}}, \cdots, \mathbf{f}_M^{\text{3D}}\}\in\mathbb{R}^{M\times C}$, we assign pseudo-labels to each point $\mathbf{f}_n^{\text{3D}}$ as follows:
\begin{equation}\label{equ:pseudo_label}
\mathbf{f}^{\hat{t}} = \mathop{\arg\max}_{l} \psi(\mathbf{f}_n^{\text{3D}}, \mathbf{f}^t_l),
\end{equation}
where $\mathbf{f}^t_l$ is CLIP's text embedding of the $l$-th class name with a pre-defined template. $\psi$ is the dot product operation.
Furthermore, in the self-distillation stage, we also incorporate 3D geometric priors to enhance the semantic consistency of the predictions.

However, the research findings from self-supervised methods \cite{he2020momentum, chen2020simple} have highlighted that directly using the network's predictions as supervision signals can lead to model collapse due to overconfidence.
To address this issue, we leverage the predictions from the exponential moving average (EMA) model as supervision signals to train the 3D point cloud network, aiming to enhance stability and mitigate the impact of overconfidence. Specifically, we use the features extracted by the EMA model to calculate the similarity with the text feature according to Equation \ref{equ:pseudo_label}, resulting in generating a point-text pair for each point cloud.

Although using EMA model can prevent mode collapse, it is still inevitable to have noise in the prediction results. 
To mitigate this issue, we incorporate the semantic consistency of 3D geometric priors during the self-distillation process. 
Specifically, we implement a voting mechanism within each superpoint to refine the prediction results. 
After assigning the most similar textual feature to each point cloud according to Equation \ref{equ:pseudo_label}, we calculate the proportion of textual features within each superpoint. 
We then modify the text corresponding to all point clouds within that superpoint to match the textual feature with the highest proportion.
This correction process effectively mitigates the impact of noise and improves the accuracy of prediction.
The voting outcome generates point-language pairs for each point cloud, then we can pull paired point and language features together while pushing unmatched features away through contrastive loss. The geometry guided self-distillation loss is as follows:
{\setlength\abovedisplayskip{3.0pt}
\setlength\belowdisplayskip{3.1pt}
\begin{equation}\label{eq:loss_contrastive}
    \mathcal{L}_{sd}=- \log \frac{\exp(\mathbf{f}^{\text{3D}}\cdot\mathbf{f}^{\hat{t}}/\tau)}{\sum_{i=1}^{n_t} \exp(\mathbf{f}^{\text{3D}}\cdot\mathbf{f}^t_i/\tau)},
\end{equation}
}where $\mathbf{f}^{\hat{t}}$  represents the textual features corresponding to the point cloud $\mathbf{f}^{\text{3D}}$; $n_t$ is the number of point-language pairs and $\tau$ is a temperature factor.


\subsection{Training and Inference}
\label{sec:inference}

\noindent\textbf{Training.} 
The training process comprises two stages. In the initial stage, we employ geometry guided distillation to train the 3D network for 70 epochs, enabling it to acquire knowledge from the 2D model. Then, we incorporate the GGSD module into an additional 30 training epochs. The total number of training epochs is consistent with OpenScene.

\noindent\textbf{Inference.} 
Using any per-point feature $\mathbf{f}^\text{3D}$ and CLIP text features obtained from a set of text prompts, we can estimate their similarities by calculating the cosine similarity score between them. This similarity score is utilized in all of our scene understanding tasks.
In all experiments, unless otherwise stated, we exclusively use pure 3D point clouds for predictions during the inference stage, without employing the 2D-3D ensemble strategy.

\section{Experiments}
\label{sec:experiments}
\subsection{Experiment Setup}
\noindent\textbf{Datasets.}
To evaluate the effectiveness of our method, we conduct experiments on three widely used public benchmarks: ScanNet~\cite{dai2017scannet}, Matterport~\cite{chang2017matterport3d}, and nuScenes~\cite{caesar2020nuscenes}. The first two datasets provide RGBD images and 3D meshes of indoor scenes, while the last one offers LIDAR scans of outdoor scenes. Additionally, Matterport3D is a challenging dataset with intricately detailed scenes, which allows us to thoroughly examine the performance of open-vocabulary queries across different datasets.

\noindent\textbf{Implementation Details.} 
We adopt the settings outlined in OpenScene~\cite{peng2023openscene}, utilize MinkowskiNet18A~\cite{choy20194d} as the 3D backbone network, and employ a simple prompt engineering technique to extract CLIP text features. Specifically, we modify the text prompts for each object class ``XX'' to ``a XX in a scene'', such as ``a chair in a scene''.
On the ScanNet and Matterport3D datasets, we employ a voxel size of 2cm with MinkowskiNet while utilizing a voxel size of 5cm on the nuScenes dataset.
Based on the findings of OpenScene, we employ LSeg~\cite{li2022languagedriven} for indoor datasets and OpenSeg~\cite{ghiasi2022scaling} for outdoor dataset as the 2D teacher model.
The temperature $\tau$ in Eq. \ref{eq:loss_contrastive} is empirically set as 0.01. 
We utilize the Adam optimizer with an initial learning rate of $1e{-}4$ for the entire training process. 
We use a batch size of 8 on ScanNet and Matterport3D with a single NVIDIA A100 GPU (80G). On nuScenes, we use a batch size of 16 with 4 A100 GPUs.
It takes around 24 hours to train and 0.1 second for inference.
More details can be found in the \textbf{supplementary file}.

\subsection{Main Results}
\label{label:s__main_results}
We conducted a series of experiments, including traditional closed-set scenarios, in-domain open vocabulary scenarios, and out-of-domain open vocabulary scenarios to test the effectiveness of the proposed methods in various 3D scene understanding tasks.

\begin{table}[!t]
    \begin{center}
    \footnotesize
    \setlength{\tabcolsep}{0.2cm}
    \resizebox{0.6\columnwidth}{!}{
    \begin{tabular}{l|cc|cc}
    \toprule
    {}  & \multicolumn{2}{c|}{ScanNet~\cite{dai2017scannet}}  & \multicolumn{2}{c}{nuScenes~\cite{caesar2020nuscenes}}\\
    \midrule
    {}   &mIoU & mAcc& mIoU & mAcc\\\hline
    \multicolumn{4}{l}{\textit{Fully-supervised methods}} \\
    TangentConv~\cite{tatarchenko2018tangent} &40.9 & - &- & -  \\ 
    TextureNet~\cite{huang2019texturenet} &54.8 & - &- & -  \\
    ScanComplete~\cite{dai2018scancomplete} & 56.6& - &- & - \\
    DCM-Net~\cite{schult2020dualconvmesh} &65.8& - &- & -  \\
    Mix3D~\cite{nekrasov2021mix3d} &{73.6} & - & - & - \\
    VMNet~\cite{hu2021vmnet} &73.2 & - &- & -  \\
    LidarMultiNet~\cite{ye2023lidarmultinet} & - & - & {82.0} & -  \\ 
    MinkowskiNet~\cite{choy20194d} & 69.0 & 77.5 & 78.0 & 83.7 \\ \hline
    \multicolumn{4}{l}{\textit{Open vocabulary methods}} \\
     \rowcolor{LightGrey}
    MSeg~\cite{lambert2020mseg} Voting & 45.6 & 54.4 & 31.0 & 36.9  \\
    \rowcolor{LightGrey}
    PLA~\cite{ding2023pla} & 17.7 & 33.5 & - & -   \\
     \rowcolor{LightGrey}
    CLIP2Scene~\cite{chen2023clip2scene}  & 25.1 & - & 20.8 & -  \\
    \rowcolor{LightGrey}
    CNS~\cite{chen2023towards} & 26.8 & - & 33.5 & -  \\
    \rowcolor{LightGrey}
    CLIP-FO3D~\cite{zhang2023clip} & 30.2 & 49.1 & - & -    \\
    \rowcolor{LightGrey}
    OpenScene$^\dagger$~\cite{peng2023openscene}  & 54.2 & 66.6 & 42.1 & 61.8  \\ 
    OpenScene$^\ddagger$~\cite{peng2023openscene}  & 52.9 & 63.2 & 42.9 & 57.1  \\ 
    \textbf{GGSD (Ours)}  & \textbf{56.5} & \textbf{68.6} & \textbf{46.1} & \textbf{59.2}  \\ 
    \bottomrule
\end{tabular}}
\end{center}
    \caption{
        \textbf{Comparisons on Standard Benchmarks.}
The numbers in gray are obtained from previous papers. 
The symbol $^\dagger$ indicates the use of the 2D-3D ensemble strategy, while $^\ddagger$ denotes the pure 3D results obtained from the official open-source model.
Our method surpasses the current state-of-the-art by a large margin in both indoor and outdoor scenarios without using the 2D-3D ensemble strategy.
        }
\label{tab:benchmark}
\end{table}

\noindent\textbf{Comparison on Standard Benchmarks.}
We compare our approach with both fully-supervised and open vocabulary methods on all classes of the ScanNet~\cite{dai2017scannet} validation set and nuScenes~\cite{caesar2020nuscenes} validation set.
As shown in the bottom panel of Table \ref{tab:benchmark}, our approach, GGSD, consistently outperforms recent open vocabulary methods PLA~\cite{ding2023pla}, CLIP2Scene~\cite{chen2023clip2scene}, CNS~\cite{chen2023towards} and CLIP-FO3D~\cite{zhang2023clip} by a large margin on both datasets. Additionally, our method exhibits significant performance improvements compared to the baseline model OpenScene~\cite{peng2023openscene}. Specifically, on the ScanNet dataset, GGSD achieves a 3.6\% increase in mIoU and a 5.4\% increase in mACC. On the nuScenes dataset, GGSD demonstrates a 3.2\% increase in mIoU and a 2.1\% increase in mAcc. 
Except for OpenScene, the other methods in Table \ref{tab:benchmark} do not employ the 2D-3D ensemble strategy due to the significant computational and storage costs associated with maintaining a 2D segmentation network and processing a large number of images per scene during testing.
It is worth noting that our method using only point clouds still outperforms OpenScene with the ensemble strategy.
Given that our method is evaluated on both indoor and outdoor scenes, and it even surpasses OpenScene with ensemble strategy, these results clearly demonstrate the effectiveness and superiority of GGSD.

\noindent\textbf{Comparison on Open-Vocabulary Scenarios.}
The standard ScanNet benchmark only contains a small vocabulary of 20 classes. To evaluate the open vocabulary capability of GGSD, we expand the original vocabulary size by using the NYU-40 label set. We remove NYU-40 labels that lack specific semantics, such as ``other structure'', ``other furniture'', and ``other prop''. The remaining categories are evenly divided into \textit{Head}, \textit{Common}, and \textit{Tail} based on the number of samples in each category (see \textbf{supplementary file} for specific division). We present the quantitative results on these three categories in Table \ref{tab:ov}.

\begin{table}[!ht]
\begin{center}
\footnotesize
\setlength\tabcolsep{3.5pt}
 \resizebox{0.67\columnwidth}{!}{
  \begin{tabular}{l|cc|cc|cc}
    \toprule
     & \multicolumn{2}{c}{Head} & \multicolumn{2}{c}{Common} & \multicolumn{2}{c}{Tail} \\
     & mIoU  & mAcc & mIoU  & mAcc & mIoU  & mAcc \\
    \midrule
    \multicolumn{6}{l}{\textit{Inference by feature projection}} \\
    \rowcolor{LightGrey}
    MaskCLIP-3D  & 19.8 & 33.1 & 13.3 & 26.8 & 7.5 & 15.3 \\
    \midrule
    OpenScene$^\ddagger$~\cite{peng2023openscene} & {64.4} & {76.8} & 27.8 & 35.1 & 14.9 & 21.4 \\
    \textbf{GGSD (Ours)} & \textbf{66.8} & \textbf{78.8} & \textbf{30.1} & \textbf{40.9} & \textbf{16.0} & \textbf{24.2} \\

    \bottomrule
  \end{tabular}}
  \end{center}
  \caption{\textbf{Comparison on open-vocabulary scenarios.} Open-vocabulary semantic segmentation on ScanNet with extended labels from the NYU label set. 
  Categories are divided into \textit{Head}, \textit{Common} and \textit{Tail} by point numbers. The numbers in gray are obtained from previous papers. 
The symbol $^\ddagger$ denotes the pure 3D point results obtained from the official open-source model.}
\label{tab:ov}
\end{table}

\begin{table}[!ht]
\begin{center}
\footnotesize
\setlength\tabcolsep{3.5pt}
 \resizebox{0.85\columnwidth}{!}{
  \begin{tabular}{l|cc|cc|cc|cc}
    \toprule
     & \multicolumn{2}{c}{Matterport21} & \multicolumn{2}{c}{Matterport40} & \multicolumn{2}{c}{Matterport80} & \multicolumn{2}{c}{Matterport160}\\
     & mIoU  & mAcc & mIoU  & mAcc & mIoU  & mAcc & mIoU  & mAcc \\     
    \midrule
    {OpenScene$^\ddagger$~\cite{peng2023openscene}} &36.0 & 48.0  &21.1 &27.5  &10.8 &13.9 &6.0 &8.1\\
{\textbf{GGSD (Ours)}}  &\textbf{40.1}  &\textbf{54.4} & \textbf{22.8} & \textbf{31.6} & \textbf{11.9} & \textbf{16.2}  & \textbf{6.3} & \textbf{9.6}\\
    \bottomrule
  \end{tabular}}
  \end{center}
 \caption{
        \textbf{Comparison on cross-domain scenarios.}
        Both GGSD and OpenScene are trained on ScanNet, and zero-shot tested on the Matterport3D dataset. 
The symbol $^\ddagger$ denotes the pure 3D results obtained from the official open-source model. $K=21$ is derived from the original Matterport3D benchmark. For $K=40, 80, 160$, we use the $K$ most common categories from the NYU label set provided in the benchmark.
        }
\label{tab:cross-domain}
\end{table}

GGSD achieves superior results over OpenScene in all categories, with a specific improvement of 2.4\% and 2.3\% in mIOU for the \textit{Head} and \textit{Common} categories, respectively. Since the categories in the \textit{Common} set typically consist of objects with smaller sizes (\eg, night stand and floor mat), the 2.3\% mIOU improvement demonstrates the stronger open vocabulary capability of our method.
However, for the \textit{Tail} category, although GGSD surpasses OpenScene by 2.8\% in mACC, the performance of both OpenScene and GGSD is not satisfactory. This can be attributed to the small sizes and the limited number of samples in these categories (\eg, lamp, box, bag, and book).

\noindent\textbf{Comparison on Cross-Domain Scenarios.}
Our GGSD method has demonstrated excellent potential in solving open vocabulary scene understanding tasks within a specific domain. However, the exploration of open vocabulary learners that can transfer across datasets is also worth investigation, as they face the dual challenges of category and data distribution.
We conduct zero-shot domain transfer experiments by training the model on ScanNet~\cite{dai2017scannet} and testing it on the Matterport3D~\cite{chang2017matterport3d} dataset without fine-tuning. It should be noted that as $K$ increases, the test set will contain more small-sized objects.
As shown in Table~\ref{tab:cross-domain}, our GGSD method consistently outperforms OpenScene in Matterport21, Matterport40, Matterport80, and Matterport160, demonstrating its strong open vocabulary capability and generalization ability.

\begin{table}[!ht]
    \begin{center}
    \footnotesize
    \setlength{\tabcolsep}{0.2cm}
    \resizebox{0.75\columnwidth}{!}{
    \begin{tabular}{l|cc|cc}
    \toprule
    {}  & \multicolumn{2}{c|}{ScanNet~\cite{dai2017scannet}}  & \multicolumn{2}{c}{Matterport3D~\cite{chang2017matterport3d}}\\
    \midrule
    {}   &mIoU & mAcc& mIoU & mAcc\\\hline
    2D Fusion Projection  & 50.0 & 62.7 & 32.3 & 40.0\\
    Pixel-Point Distillation & 52.9 & 63.2 & 36.1 & 48.0\\
    Geometry Guided Distillation & 53.5 & 65.0 & 36.7 & 49.3\\
    Self-Distillation & 56.1 & 68.2 & 39.0 & 53.3\\
    {Geometry Guided Self-distillation}  & {56.5} & {68.6} & {40.1} & {54.4} \\ 
    \bottomrule
\end{tabular}}
\end{center}
    \caption{
        \textbf{Ablation study of GGSD.}
        Comparison of semantic segmentation performance of each component of our GGSD method.
        }
\label{tab:ablation_component}
\end{table}

\subsection{Ablation Studies and Analysis}
In this section, we investigate our key components through an in-depth ablation study. The experiments are conducted on the ScanNet dataset and subsequently tested on both ScanNet and Matterport3D datasets.

\noindent\textbf{Component Analysis.}
As shown in Table \ref{tab:ablation_component}, even without utilizing 3D geometric priors, pixel-point distillation can outperform 2D direct projection, demonstrating the  advantage of 3D representation. 
By incorporating 3D geometric priors, the geometry guided distillation method further enhances performance, resulting in a 0.6\% increase in mIoU and a 1.8\% increase in mAcc on the ScanNet dataset.
Through self-distillation, we achieve a notable enhancement in performance. Specifically, on the ScanNet dataset, there is a substantial increase of 2.6\% in mIoU and 3.2\% in mAcc. On the Matterport3D dataset, the mIoU is increased by 2.3\%, and the mAcc is improved by 4.0\%.
Finally, the utilization of the voting mechanism within the superpoints leads to improved performance by mitigating the impact of noise through the refinement model's predictions.
It is noteworthy that our GGSD method demonstrates substantial improvements compared to the 2D teacher model, \ie, 2D Fusion Projection, achieving a +6.5\% mIoU and +5.9\% mAcc on the ScanNet dataset, as well as a +7.8\% mIoU and +14.4\% mAcc on the Matterport3D dataset.
These results provide strong evidence for the effectiveness of our method.

\begin{table}[!t]
\begin{center}
\footnotesize
    \setlength{\tabcolsep}{0.2cm}
    \resizebox{0.75\columnwidth}{!}{
    \begin{tabular}{l|cc|cc}
    \toprule
    {}  & \multicolumn{2}{c|}{ScanNet~\cite{dai2017scannet}}  & \multicolumn{2}{c}{Matterport3D~\cite{chang2017matterport3d}}\\
    \midrule
    {}   &mIoU & mAcc& mIoU & mAcc\\\hline
    Geometry Guided Distillation & 54.0 & 65.0 & 36.7 & 49.3\\
    \hline
    Fixed 2D model & 51.9 & 64.6 & 34.2 & 45.7\\
    Fixed 3D model & 55.0 & 65.8 & 37.3 & 49.8\\
    EMA model & 56.1 & 68.2 & 39.0 & 53.3\\
    \bottomrule
\end{tabular}}
\end{center}
    \caption{
        {Comparisons of the 3D supervision signals provided by different models.}
        }
\label{tab:model}
\end{table}

\noindent\textbf{The Effectiveness of the EMA Model.}
In the second stage, we employ the EMA model to provide supervised signals for self-distillation. In Table \ref{tab:model}, we investigate the performance of different models in providing supervised signals. When using the fixed 2D model for pixel prediction and projecting it onto points, we observe lower performance compared to geometry-guided distillation. This can be attributed to the fact that the 2D model provides more erroneous supervised signals. Furthermore, using the fixed first-stage distilled 3D model only results in marginal performance improvement. However, when utilizing the EMA model, we achieve the best performance, primarily because the EMA model gradually obtains more accurate prediction results as training progresses.

\begin{table}[t]
    \begin{center}
    \footnotesize
    \setlength{\tabcolsep}{0.2cm}
    \resizebox{0.8\columnwidth}{!}{
    \begin{tabular}{l|cc|cc}
    \toprule
    {}  & \multicolumn{2}{c|}{ScanNet~\cite{dai2017scannet}}  & \multicolumn{2}{c}{Matterport3D~\cite{chang2017matterport3d}}\\
    \midrule
    {}   &mIoU & mAcc& mIoU & mAcc\\\hline
        Pixel-Point Distillation & 52.9 & 63.2 & 36.1 & 48.0\\
    SAM \cite{kirillov2023segment} Refine & 53.0 & 63.3 & 35.9 & 48.4\\
    Geometry Guided Distillation & 53.5 & 65.0 & 36.7 & 49.3\\
    \bottomrule
\end{tabular}}
\end{center}
    \caption{
        {Comparisons between the utilization of geometric priors and SAM \cite{kirillov2023segment}.}
        }
\label{tab:sam}
\end{table}

\newcommand{\width}{0.24\textwidth}
\begin{figure*}[t]
\begin{center}        
        \renewcommand{\arraystretch}{0.5}
        \hfill{}
        \begin{tabular}{lccccc}
            \multicolumn{5}{c}{
            \includegraphics[width=0.88\textwidth]{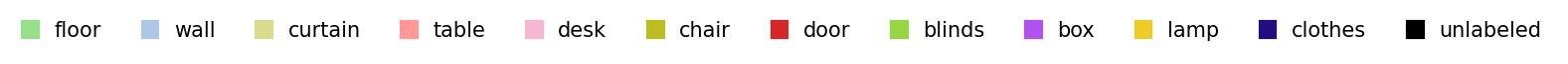}}\\
          \rotatebox{90}{\scriptsize\hspace{20pt}ScanNet~\cite{dai2017scannet}} & 
            \includegraphics[width=0.23\textwidth]{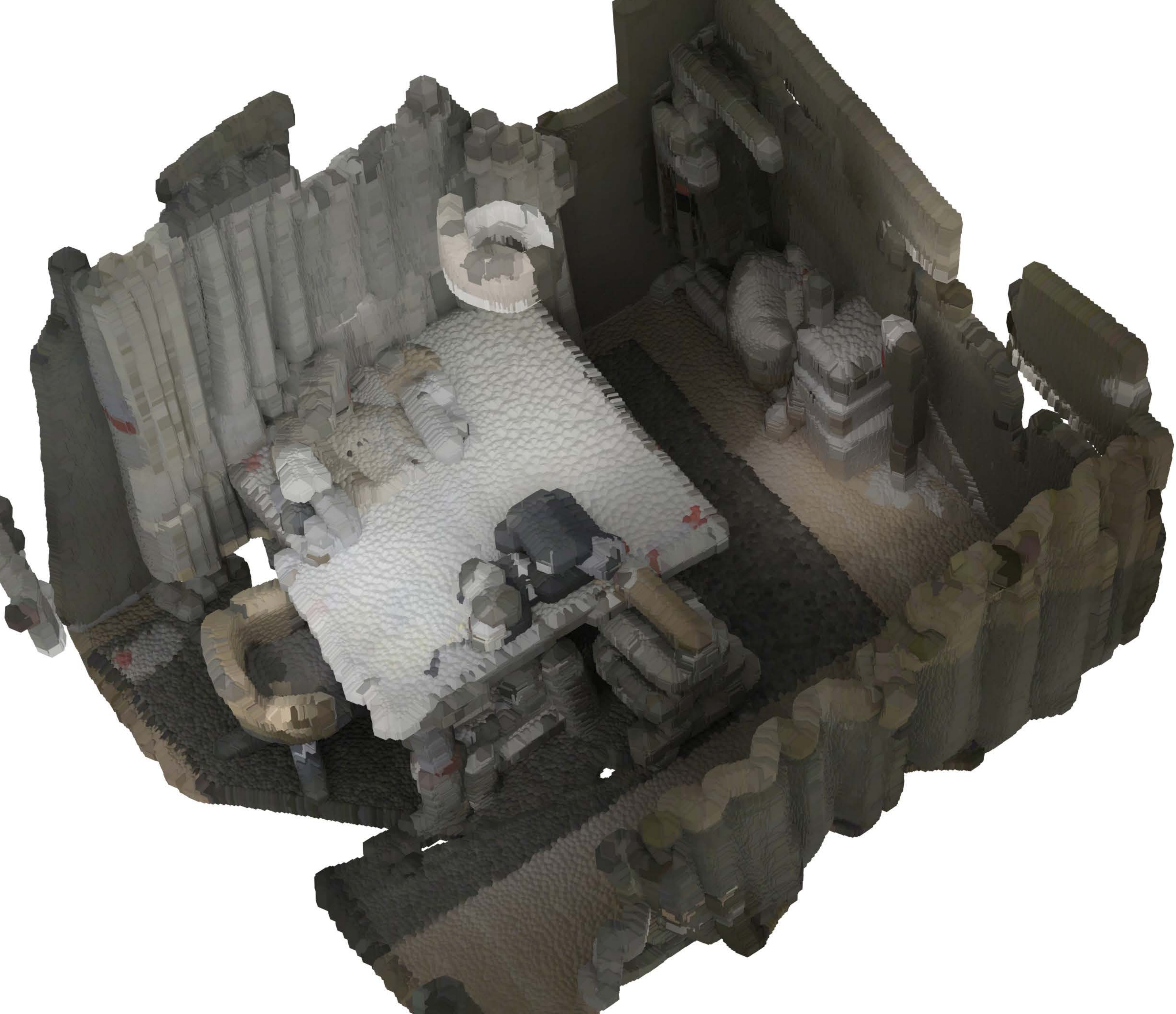}&
            \includegraphics[width=0.23\textwidth]{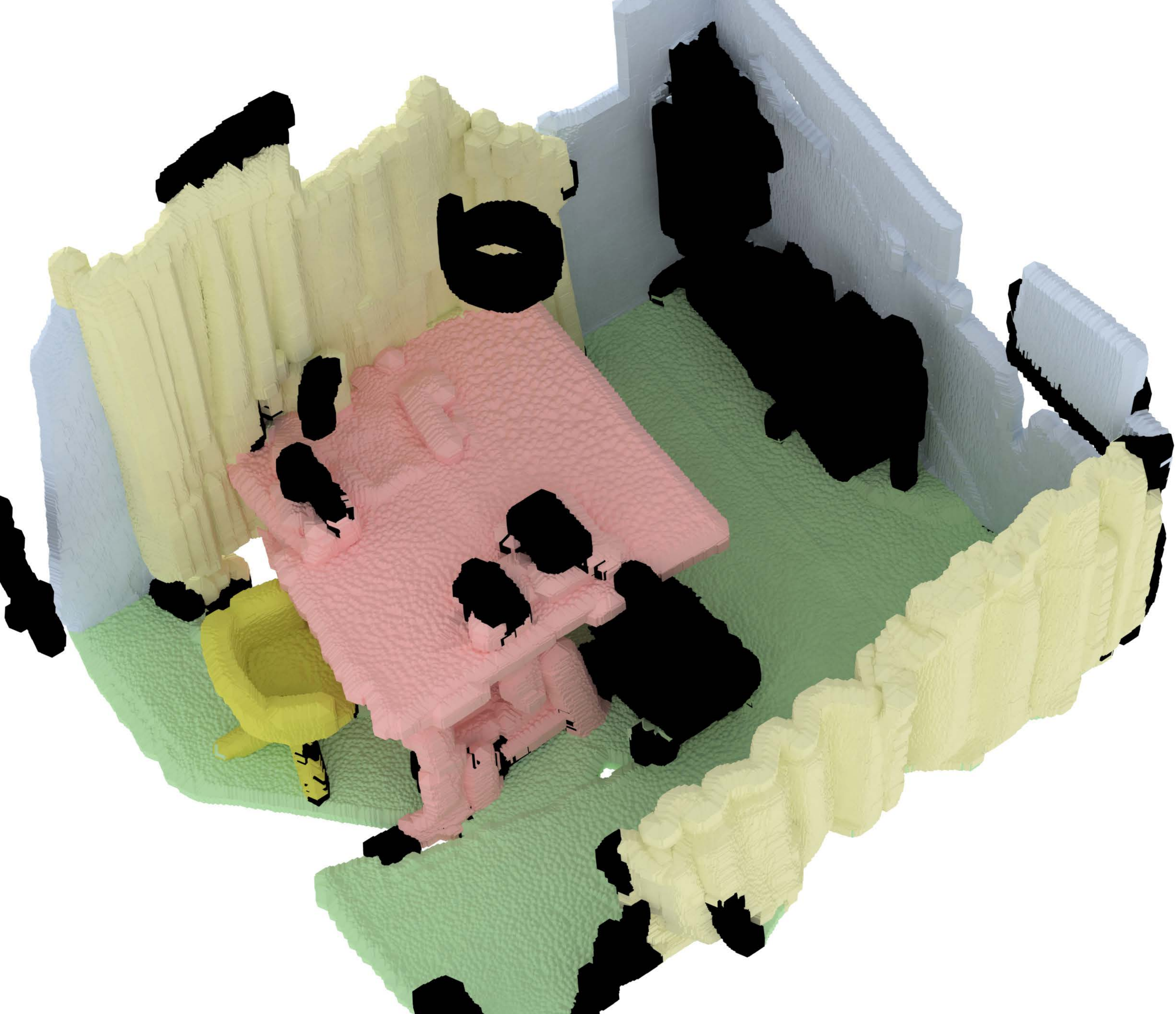}&
            \includegraphics[width=0.23\textwidth]{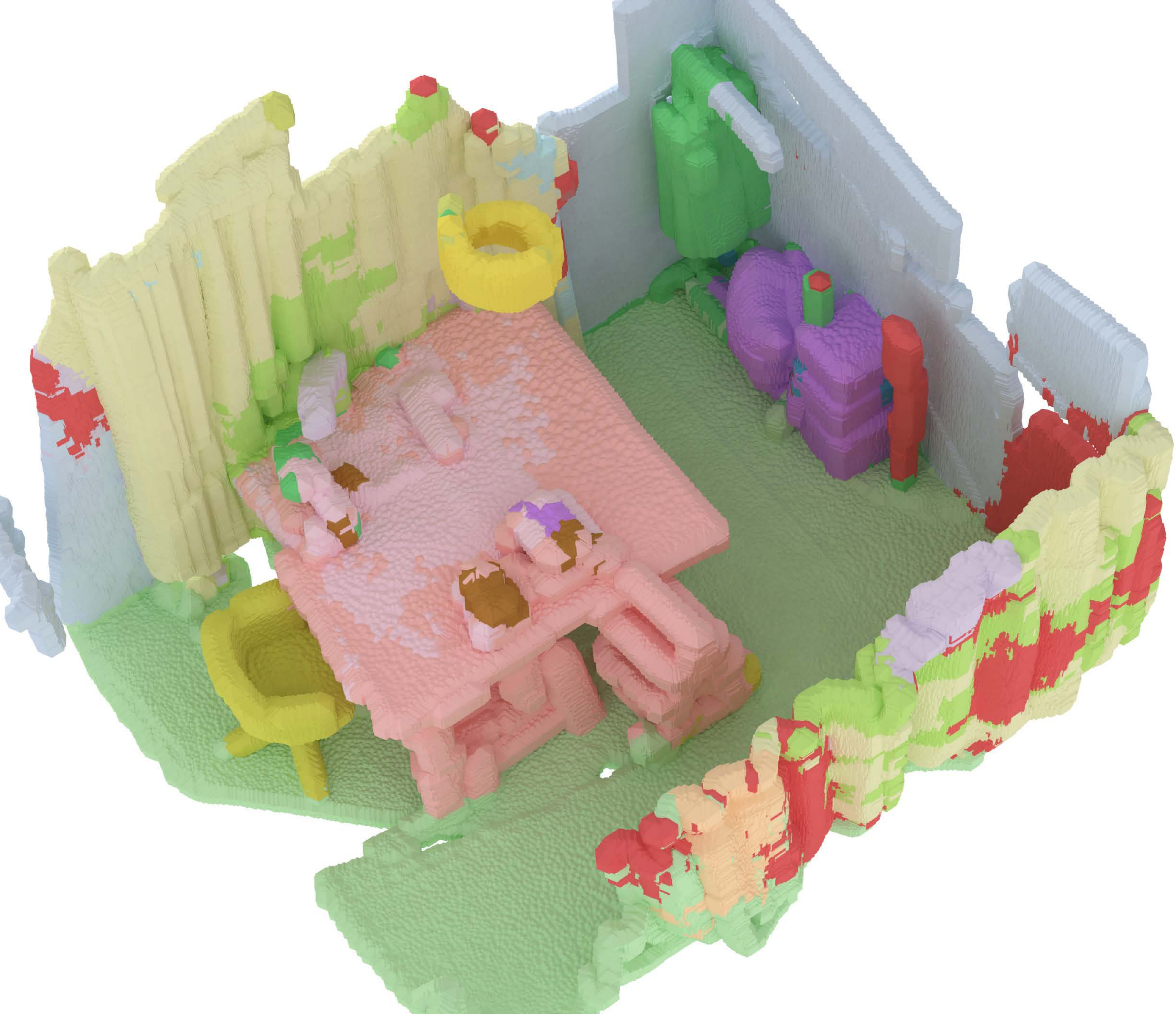}&
            \includegraphics[width=0.23\textwidth]{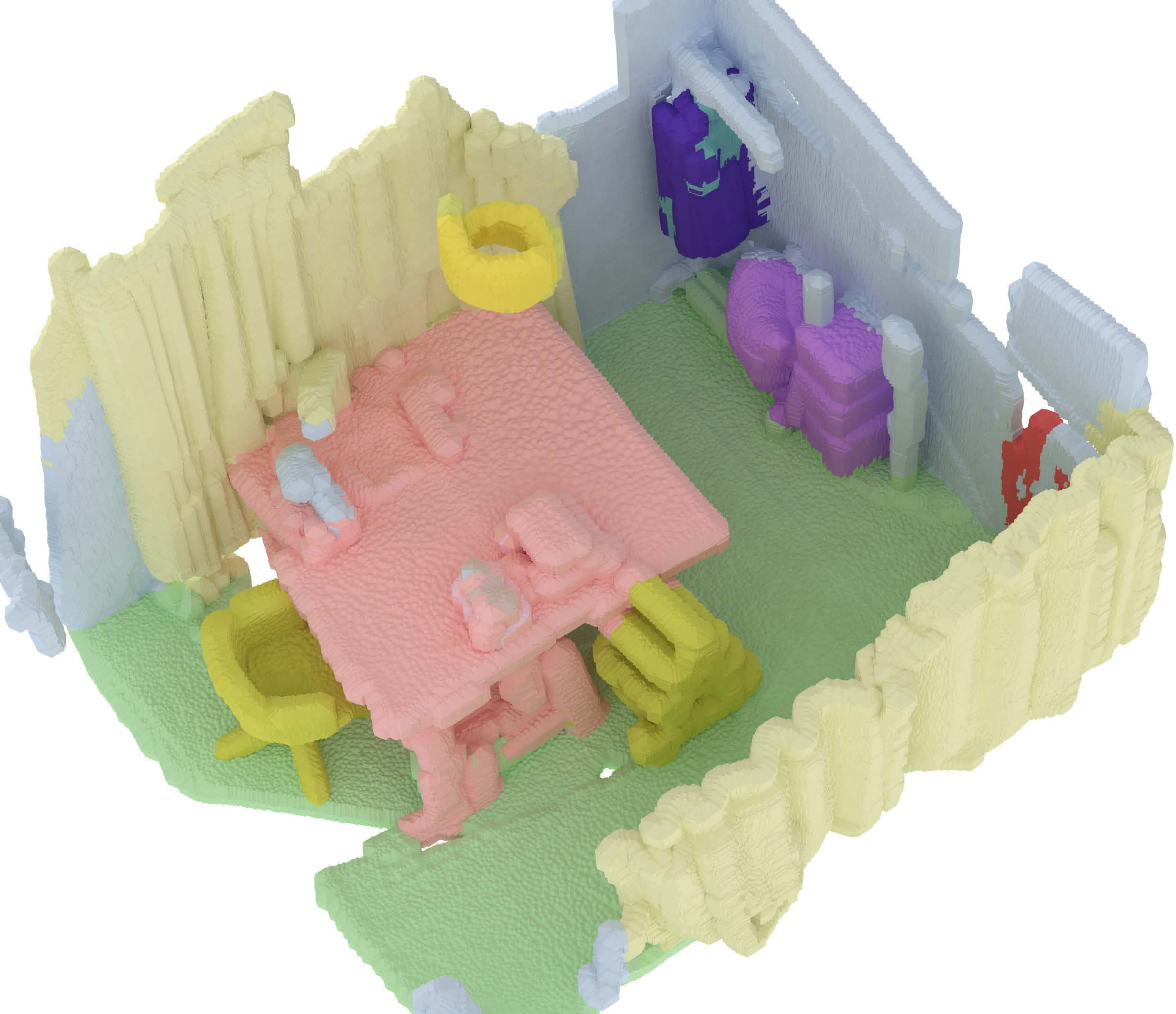}\\
            \multicolumn{5}{c}{
            \includegraphics[width=0.88\textwidth]{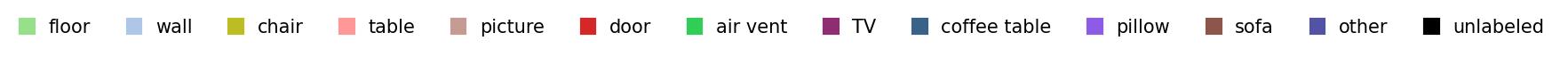}}\\
            \rotatebox{90}{\scriptsize\hspace{3pt} Matterport3D~\cite{chang2017matterport3d}} & 
            \includegraphics[width=0.23\textwidth]{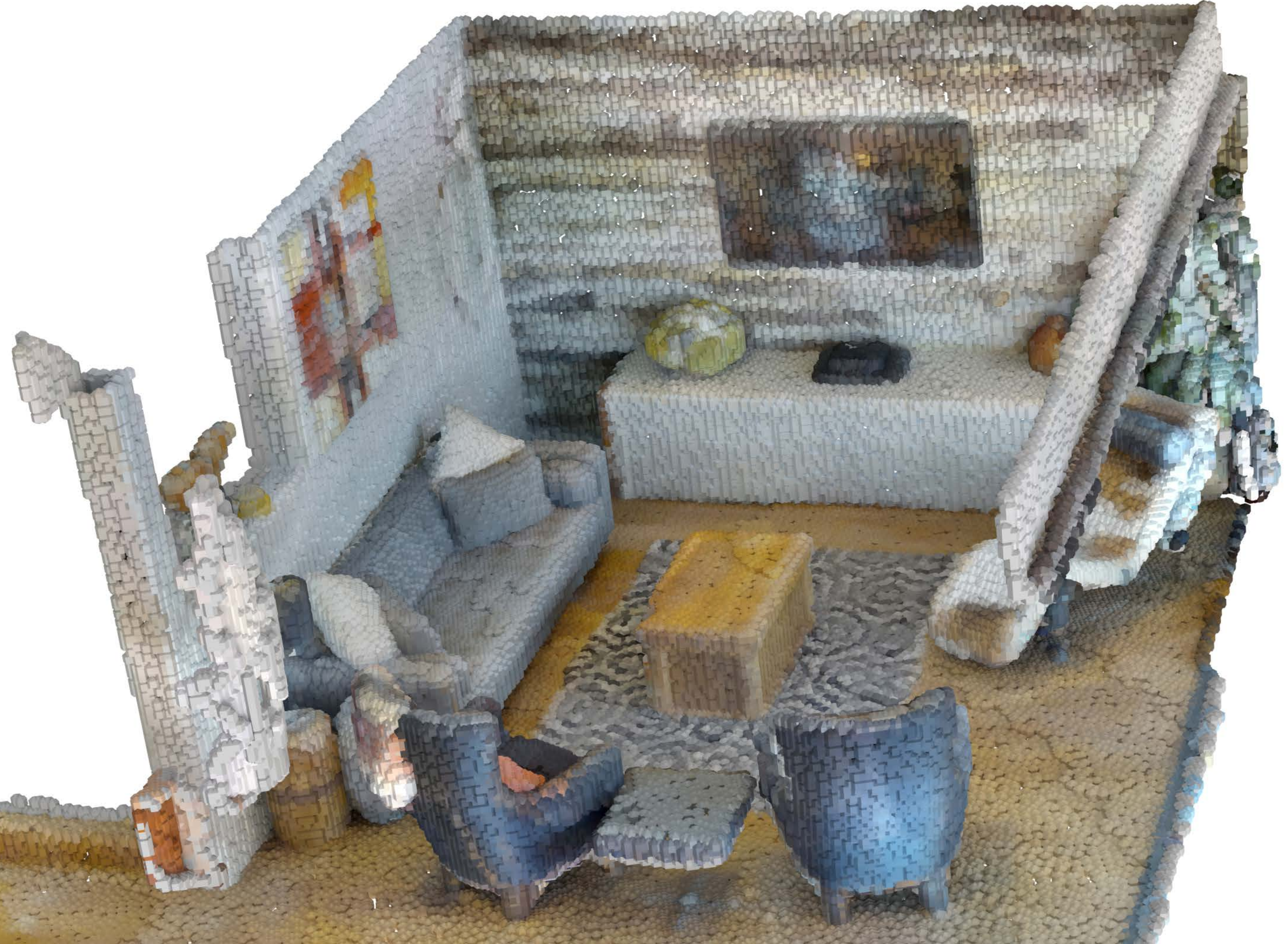}&
            \includegraphics[width=0.23\textwidth]{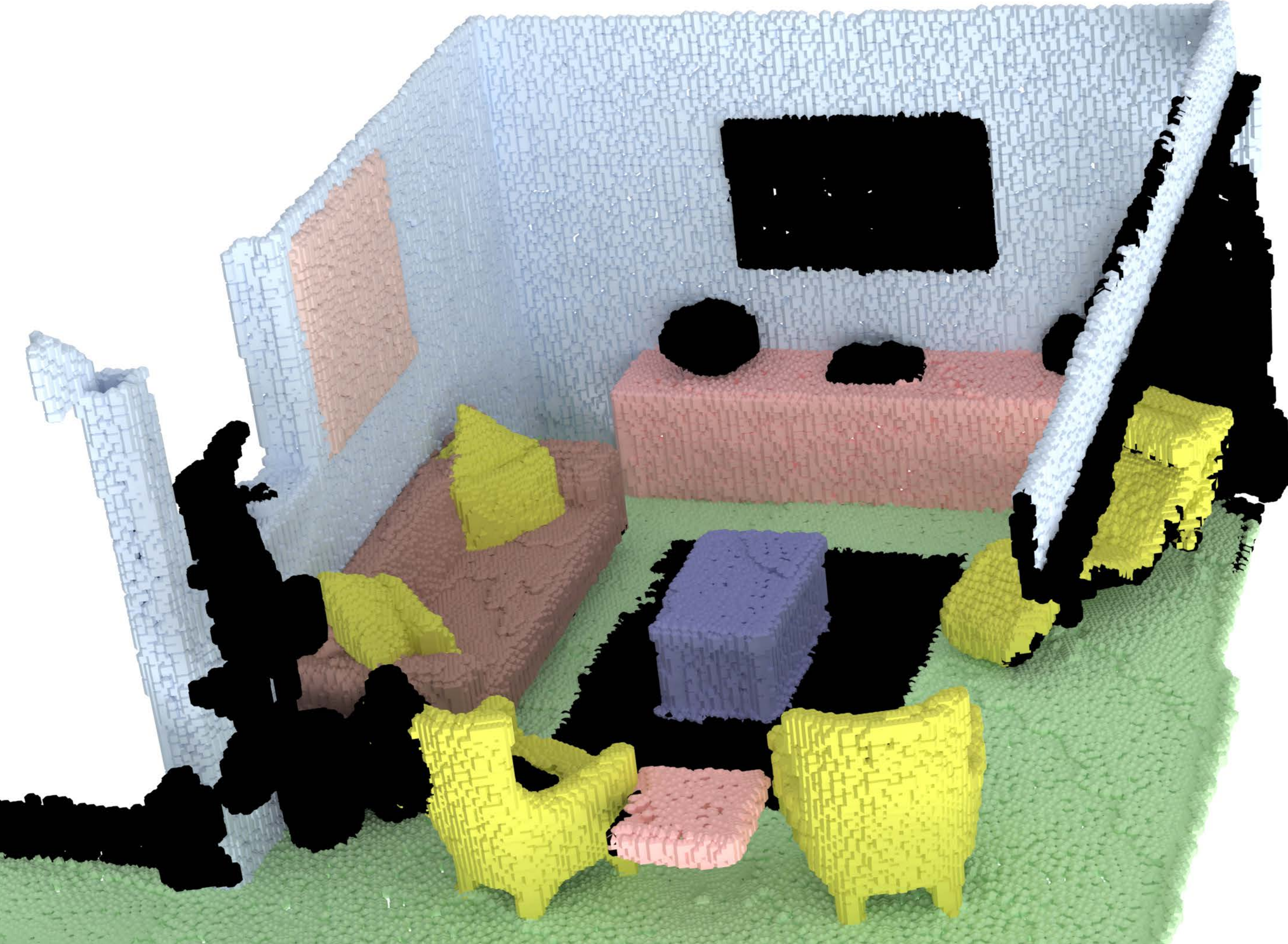}&
            \includegraphics[width=0.23\textwidth]{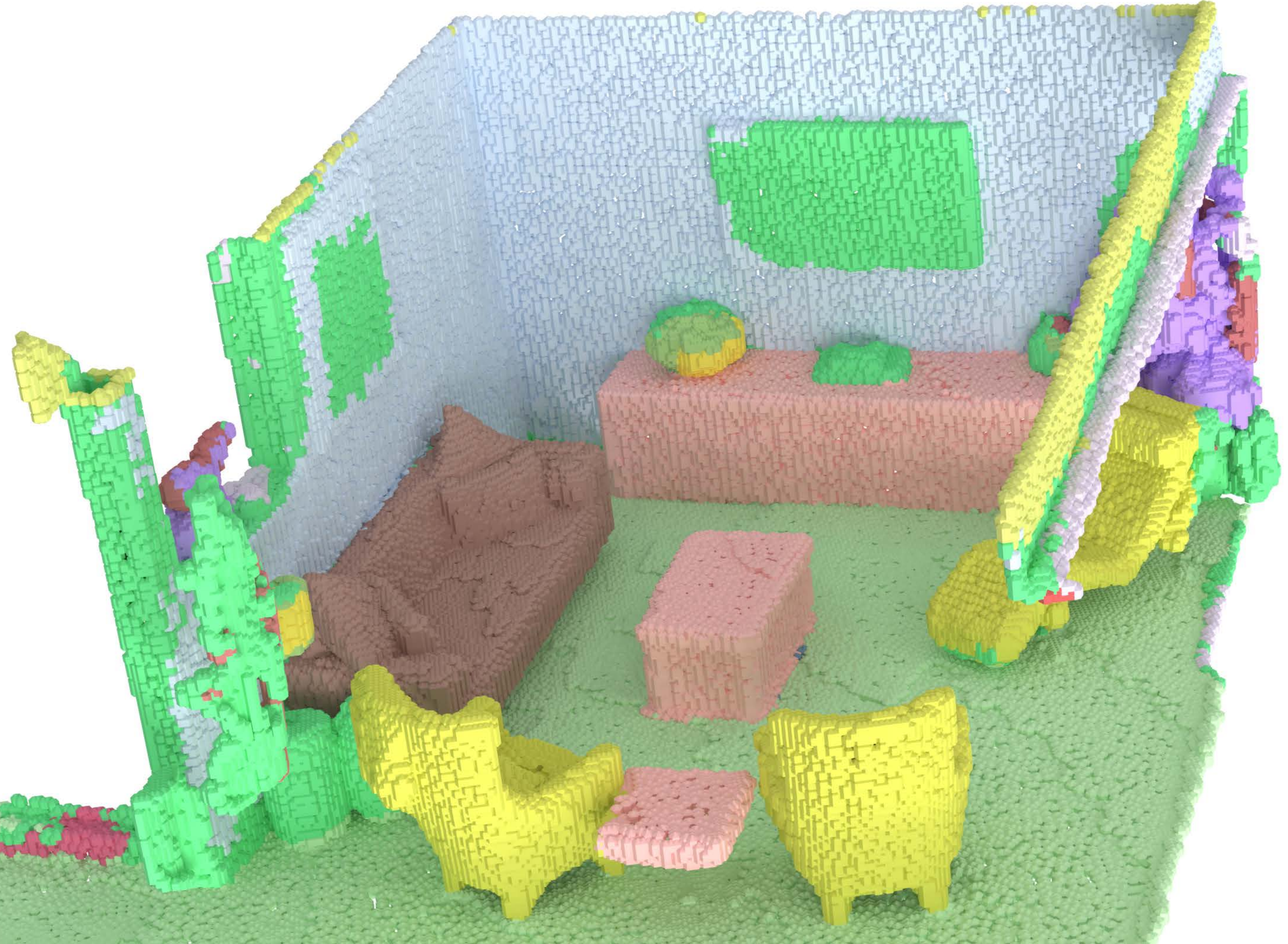}&
            \includegraphics[width=0.23\textwidth]{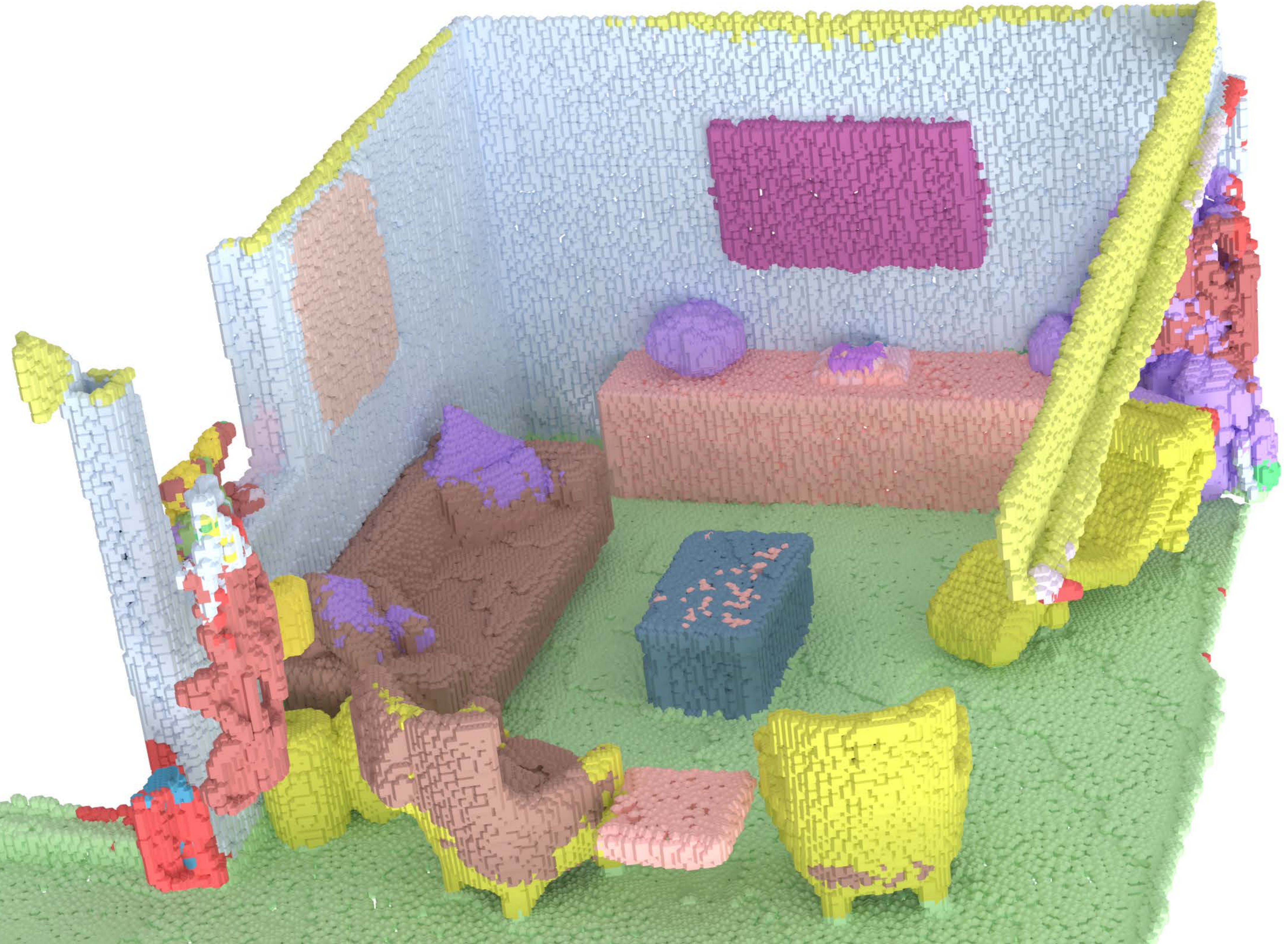}\\
            \multicolumn{5}{c}{
            \includegraphics[width=0.68\textwidth]{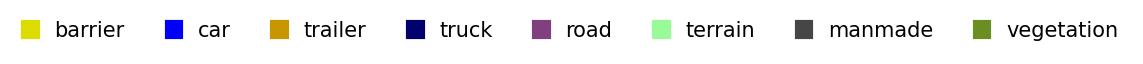}}\\
            \rotatebox{90}{\scriptsize\hspace{1pt}nuScenes~\cite{caesar2020nuscenes}} & 
            \includegraphics[width=0.21\textwidth]{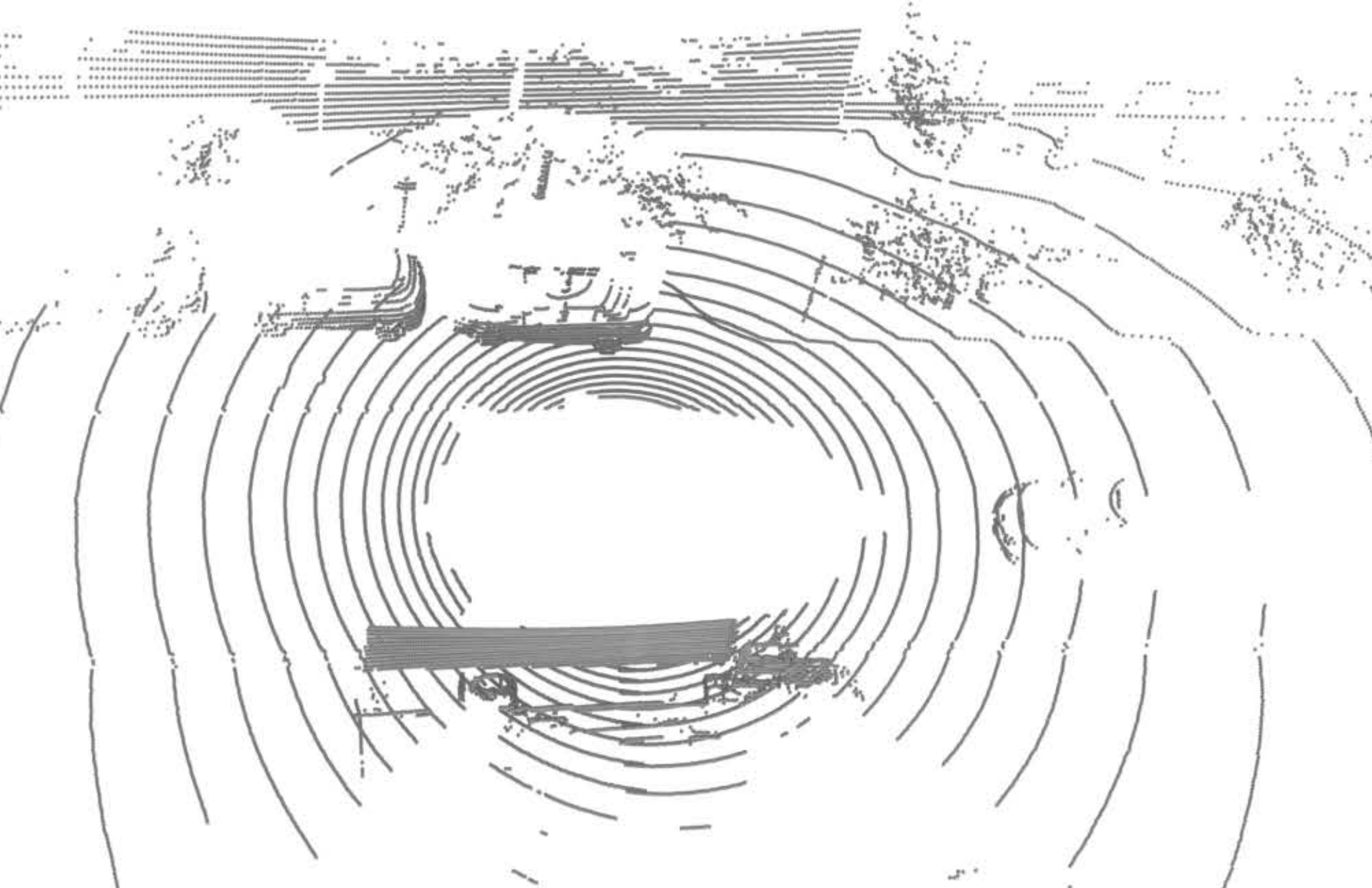}&
            \includegraphics[width=0.21\textwidth]{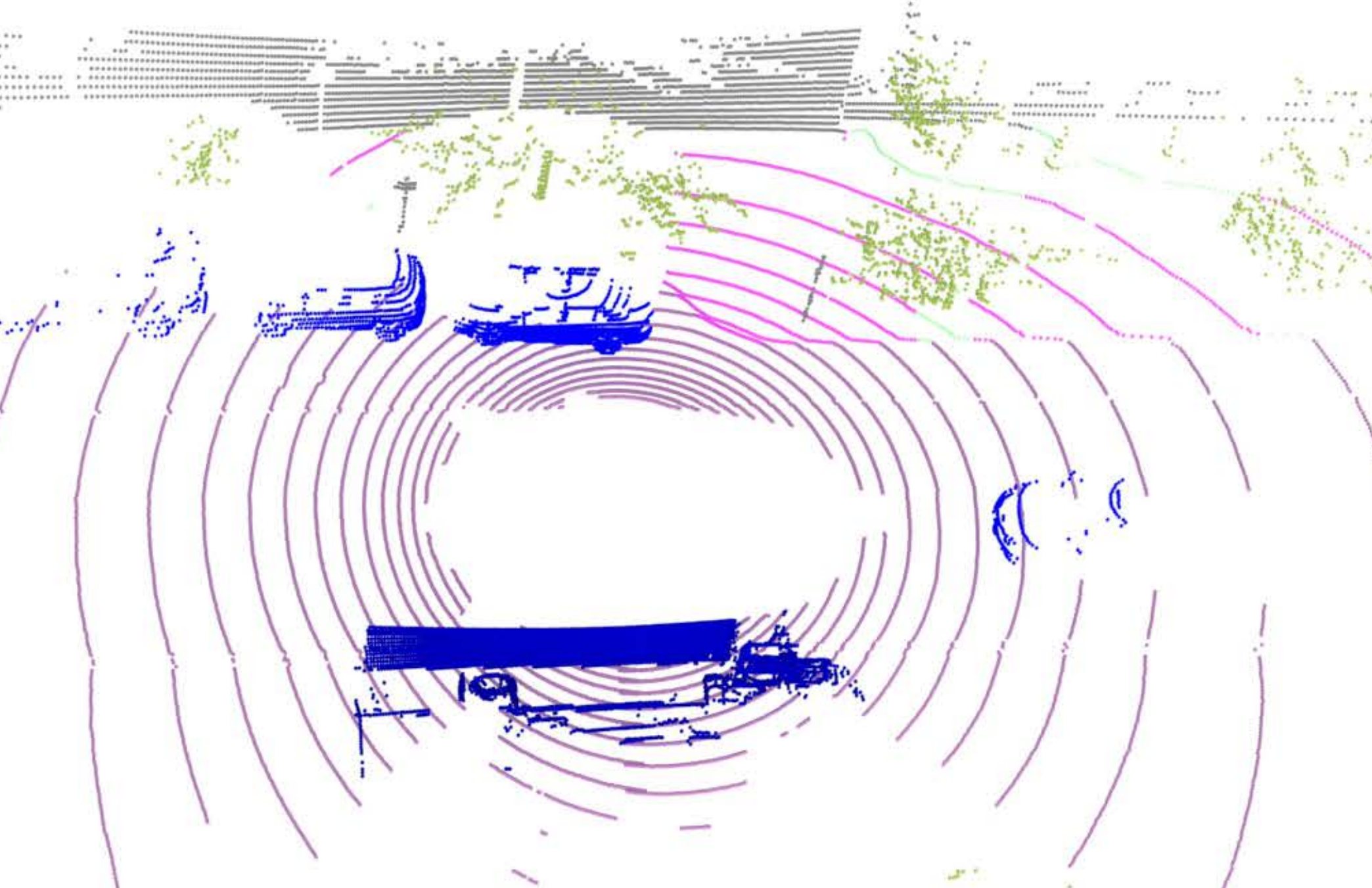}&
            \includegraphics[width=0.21\textwidth]{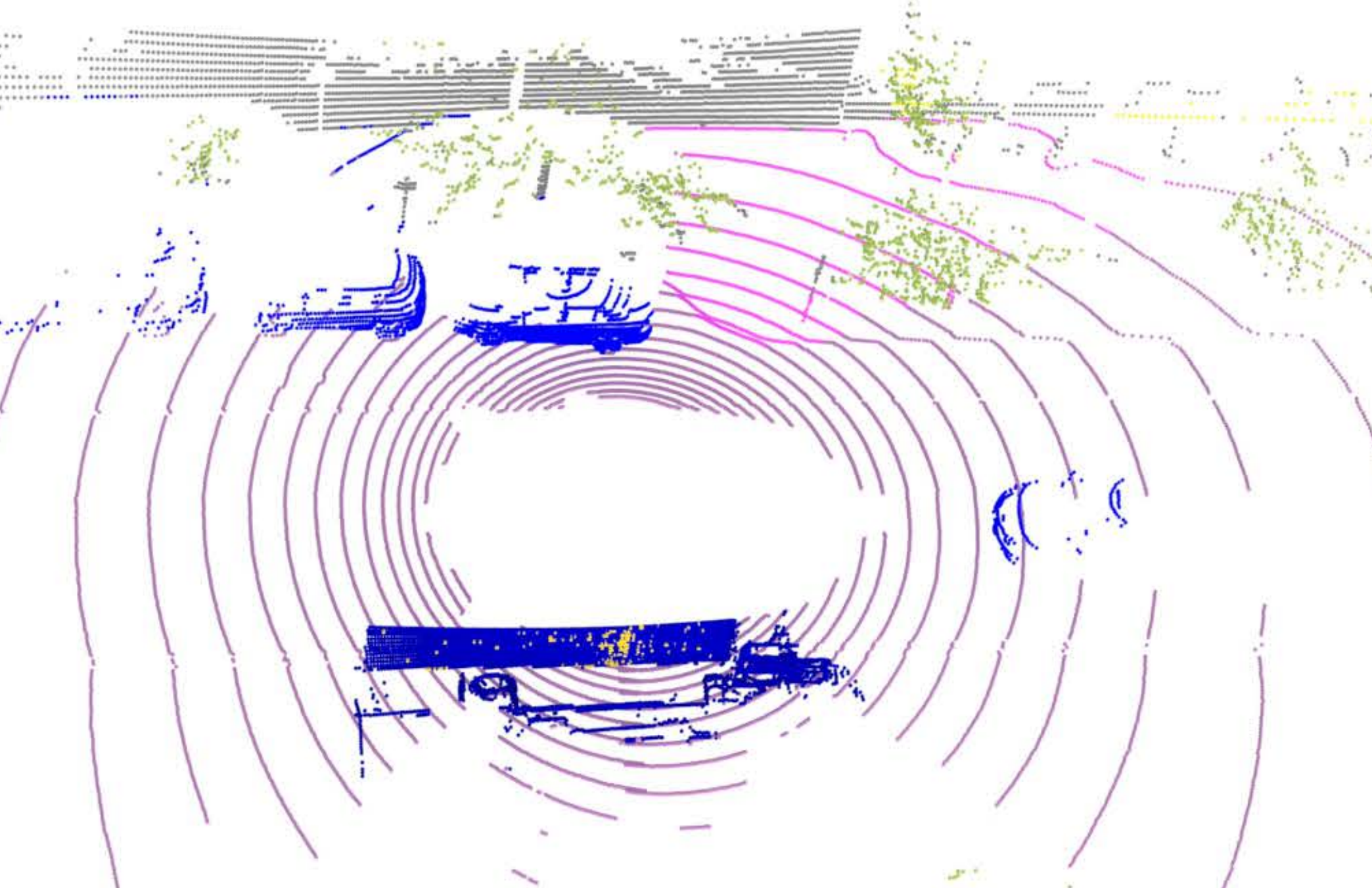}&
            \includegraphics[width=0.21\textwidth]{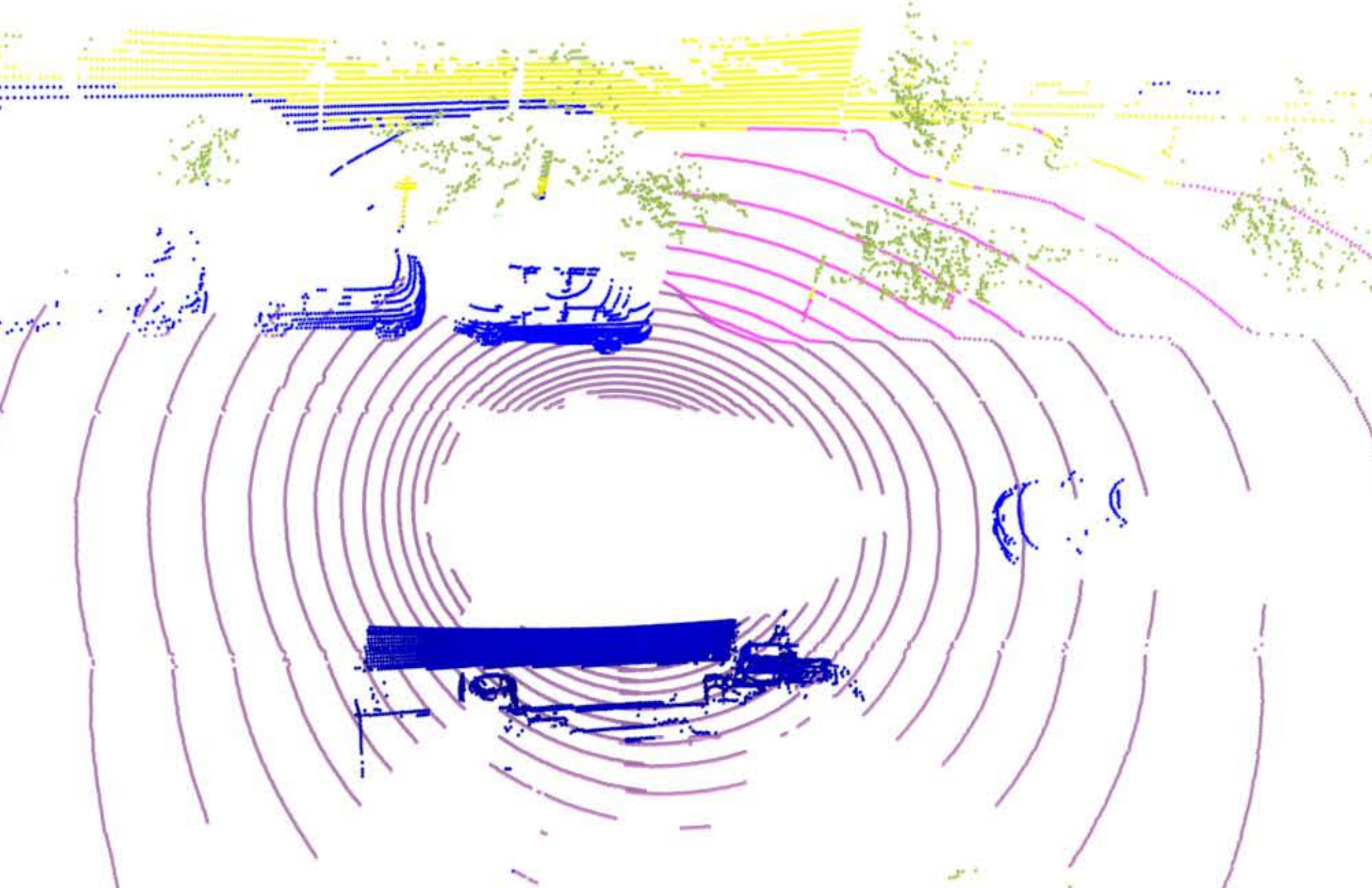}\\
            \\
              {} &{\small Input} &
              {\small GT Segmentation}  &
              {\small OpenScene~\cite{peng2023openscene}} &
              {\small \textbf{Ours}}\\
        \end{tabular}
        \end{center}
        \caption{
        \textbf{Qualitative results.} We present the qualitative results of 3D semantic segmentation on public indoor and outdoor benchmarks.
        }
        \label{fig:vis}
\end{figure*}

\noindent\textbf{The Effectiveness of 3D Geometric Priors.}
The recent work CNS~\cite{chen2023towards} utilized the foundation model SAM \cite{kirillov2023segment} to refine pseudo-labels from the 2D model, achieving performance improvements on a lower baseline (27.3\% mIoU in ScanNet). We also apply this approach to LSeg. Specifically, we follow the official implementation of SAM to generate masks for all images and perform average pooling within each mask to refine them. Finally, we utilize the extracted refined features for distillation. However, as shown in Table \ref{tab:sam}, this approach yields minimal improvements.
Conversely, leveraging 3D geometric priors results in a 0.6\% mIoU improvement in ScanNet.

\noindent\textbf{Qualitative Results.}
As shown in Fig. \ref{fig:vis}, our method successfully segments all categories that exist in closed-set ScanNet benchmarks, such as {wall} and {table}. Additionally, our method successfully segments some categories that are not annotated in traditional benchmarks, such as {box}, {lamp}, and {clothes}, demonstrating our model's excellent open vocabulary capability.
On the Matterport3D, the GGSD method successfully segments unlabeled {television} and also identifies mislabeled {pillow} and {coffee table} in the ground truth. 
Furthermore, there have been instances of segmentation failures in Matterport3D. Due to linguistic ambiguity, sofa chairs are sometimes segmented as separate entities, such as sofas and chairs, while coffee tables can be segmented as coffee tables and tables. The misclassification of certain pillow points as sofas may be attributed to their strong color resemblance to sofas.
In the case of nuScenes, GGSD also successfully segments unannotated {barrier}.
These qualitative results obtained from GGSD demonstrate its excellent open vocabulary capability. More visualization results can be found in the \textbf{supplementary file}.

\section{Conclusion}
In this study, we proposed an effective approach called Geometry Guided Self-Distillation (GGSD) to learn superior 3D representations from 2D pre-trained models. We first introduced a geometry guided distillation module that leveraged 3D geometric priors to alleviate noise in 2D models. Moreover, we harnessed the representation advantages of 3D data through geometry guided self-distillation, resulting in significant improvement in the model performance for 3D scene understanding.
Our GGSD method significantly outperformed existing open-vocabulary 3D scene understanding methods on both indoor and outdoor datasets. This highlighted the crucial role of geometric priors and of 3D data representation in distillation-based approaches for 3D scene understanding. 

\noindent\textbf{Limitations.} 
Though GGSD has achieved promising results in 3D open vocabulary scene understanding tasks, it has two main limitations. Firstly, while geometric priors can significantly improve the model performance on objects with pronounced geometric structures, the improvement can be limited on objects with fewer prominent geometric characteristics. Secondly, similar to other open vocabulary methods, our approach is susceptible to the influence of language ambiguity, as illustrated by the qualitative results in Fig. \ref{fig:vis}. 


\noindent\textbf{Acknowledgement.} This work was supported by the InnoHK program.


%
%
\bibliographystyle{splncs04}
\bibliography{main}

\end{document}


\title{ Open Vocabulary 3D Scene Understanding via Geometry Guided Self-Distillation \\- Supplementary Material -} 

\titlerunning{GGSD}

\author{Pengfei Wang\inst{1,2}\orcidlink{0000-0002-3675-9508} \and
Yuxi Wang\inst{2}\orcidlink{0000-0003-1579-2357} \and
Shuai Li\inst{1}\orcidlink{0000-0003-0760-5267}  \and
Zhaoxiang Zhang\inst{1,2,3,4}\textsuperscript{(\Letter)}  \and
Zhen Lei\inst{1,2,3,4}  \and
Lei Zhang\inst{1}\textsuperscript{(\Letter)}\orcidlink{0000-0002-2078-4215}}

\authorrunning{P. Wang et al.}

\institute{The Hong Kong Polytechnic University \and
Center for Artificial Intelligence and Robotics, HKISI, CAS 
 \and
State Key Laboratory of Multimodal Artificial Intelligence Systems, CASIA \and
School of Artificial Intelligence, University of Chinese Academy of Sciences (UCAS)
\\ 
\email{pengfei.wang@connect.polyu.hk, zhaoxiang.zhang@ia.ac.cn, cslzhang@comp.polyu.edu.hk}}

\maketitle

\renewcommand{\thetable}{A\arabic{table}}
\renewcommand{\thefigure}{A\arabic{figure}}
\numberwithin{equation}{section}

The following contents are provided in this supplementary material:
	\begin{itemize}
        \item Section \ref{sec:impl_detail}: Detailed implementations of GGSD. 
        \item Section \ref{sec:B}: More qualitative results.

        \end{itemize}

\section{Implementation Details}\label{sec:impl_detail}

\textbf{More Details of Feature Fusion.}
OpenScene can be considered as our baseline model, and we strictly follow its approach for multi-view fusion of pixel embeddings.
Specifically, on the nuScenes dataset, we perform fusion using all the images from each scene. However, on the ScanNet dataset, we sample one image out of every 20 video frames. For indoor datasets, we conduct occlusion tests. For each surface point, we first find its corresponding pixel in the image and calculate the distance between that pixel and the 3D point. Only when the difference between the distance and the depth value of that pixel is smaller than a threshold $\sigma$, do we pair the 3D point and the pixel. The threshold $\sigma$ is proportional to the depth value $D$. Due to the high noise in depth values, we use $\sigma = 0.2D$ for ScanNet and $\sigma = 0.02D$ for Matterport. We do not project the features of pixels in the ``invalid'' regions of the depth map onto 3D points. For the nuScenes LiDAR point cloud, since no depth images are provided, we do not perform occlusion tests. We only use the synchronized images and the corresponding LiDAR points at the last timestamp of a 0.5-second segment.

\textbf{More Details of Superpoint Generation.} 
We employ VCCS \cite{papon2013voxel} to generate superpoints for indoor datasets. VCCS starts with a set of seed points uniformly distributed on a voxel grid with a resolution of $R_{seed}$, and gradually expands the superpoints. Initially, we divide the input point cloud into voxel grids of size $2\times2\times2$ cm. Then, a set of seed points is uniformly distributed within the voxelized point cloud, with a spacing of 50 cm between seed points. For each seed point, within a sphere of radius 50 cm, we set the seed point as the initial center and search for its 27 neighboring points. The distance between each neighboring point and the center point is calculated using the following formula:
\begin{equation}\label{eq:vccs}
    \boldsymbol{{D}} = \sqrt{w{c}D^2{c}+\frac{w{s}D{s}}{3R^2{seed}}+w{n}D_{n}}.
\end{equation}
Here, $D{c}, D{s}, D{n}$ represent the Euclidean distances for color, spatial and normal attributes, respectively. For the hyperparameters, we directly follow the settings of GrowSP \cite{zhang2023growsp}, setting the weights $w{c}, w{s}, w{n}$ to 0.2, 0.4 and 1, respectively.

Considering that outdoor point clouds are typically dominated by ``roads'' and have significantly different point densities from indoor datasets, we adopt the random sample consensus (RANSAC) + Euclidean clustering as an alternative approach to VCCS for generating superpoints. Specifically, we utilize RANSAC to fit planes and considered points within a distance of 0.2m from the plane as a single large superpoint. After fitting the largest plane (usually corresponding to ``roads''), we employ Euclidean clustering to create superpoints for the remaining points. Specifically, if the Euclidean distance between two points is less than 0.2m, they are assigned to the same superpoint; otherwise, they are not assigned.

\begin{table}[!t]
    \centering
    \footnotesize
    \setlength{\tabcolsep}{0.25cm}
    \resizebox{0.52\textwidth}{!}{
    \begin{tabular}{l|l}
        \toprule
        nuScenes 16 labels & Our pre-defined labels\\
        \midrule
        barrier & barrier, barricade\\
        bicycle & bicycle\\
        bus & bus\\
        car  & car\\
        construction vehicle  &  \parbox[t]{4cm}{bulldozer, excavator, concrete mixer, crane, dump truck}\\
        motorcycle  & motorcycle\\
        pedestrian  & pedestrian, person\\
        traffic cone  & traffic cone\\
        trailer  & \parbox[t]{4cm}{trailer, semi trailer, cargo container, shipping container, freight container}\\
        truck  & truck\\
        driveable surface  & road\\
        other flat  & \parbox[t]{4cm}{curb, traffic island, traffic median}\\
        sidewalk  & sidewalk\\
        terrain  & \parbox[t]{4cm}{grass, grassland, lawn, meadow, turf, sod}\\
        manmade  & building, wall, pole, awning\\
        vegetation & \parbox[t]{4cm}{tree, trunk, tree trunk, bush, shrub, plant, flower, woods}\\
        \bottomrule
    \end{tabular}}
    \caption{
        \textbf{Label Mappings for nuScenes 16 Classes.} Here we list the total 43 pre-defined non-ambiguous class names corresponding to the 16 nuScenes classes. 
        }
\label{tab:nuscenes_mapping}
\end{table}

\textbf{Pre-defined Labels for nuScenes.}
For the indoor dataset, we directly utilize the predefined label names provided within the dataset. 
However, there are instances where certain class names may be ambiguous for the nuScenes benchmark. In such cases, we can take the approach employed by OpenScene and define unambiguous class names for each category in advance. To facilitate this mapping process, Table \ref{tab:nuscenes_mapping} presents the predefined class names and their corresponding nuScenes categories.  By mapping the predictions back to these 16 categories, we can have a clearer understanding of our method's performance in the nuScenes benchmark. 


\newcommand{\width}{0.23\textwidth}
\begin{figure*}[h]
        \centering
        \renewcommand{\arraystretch}{0.5}
        \hfill{}
        \begin{tabular}{lcccc}
            \multicolumn{5}{c}{
            \includegraphics[width=0.9\textwidth]{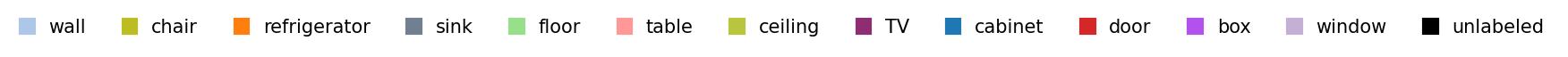}}\\
            \rotatebox{90}{\scriptsize\hspace{20pt} (a)} & 
            \includegraphics[width=0.3\textwidth]{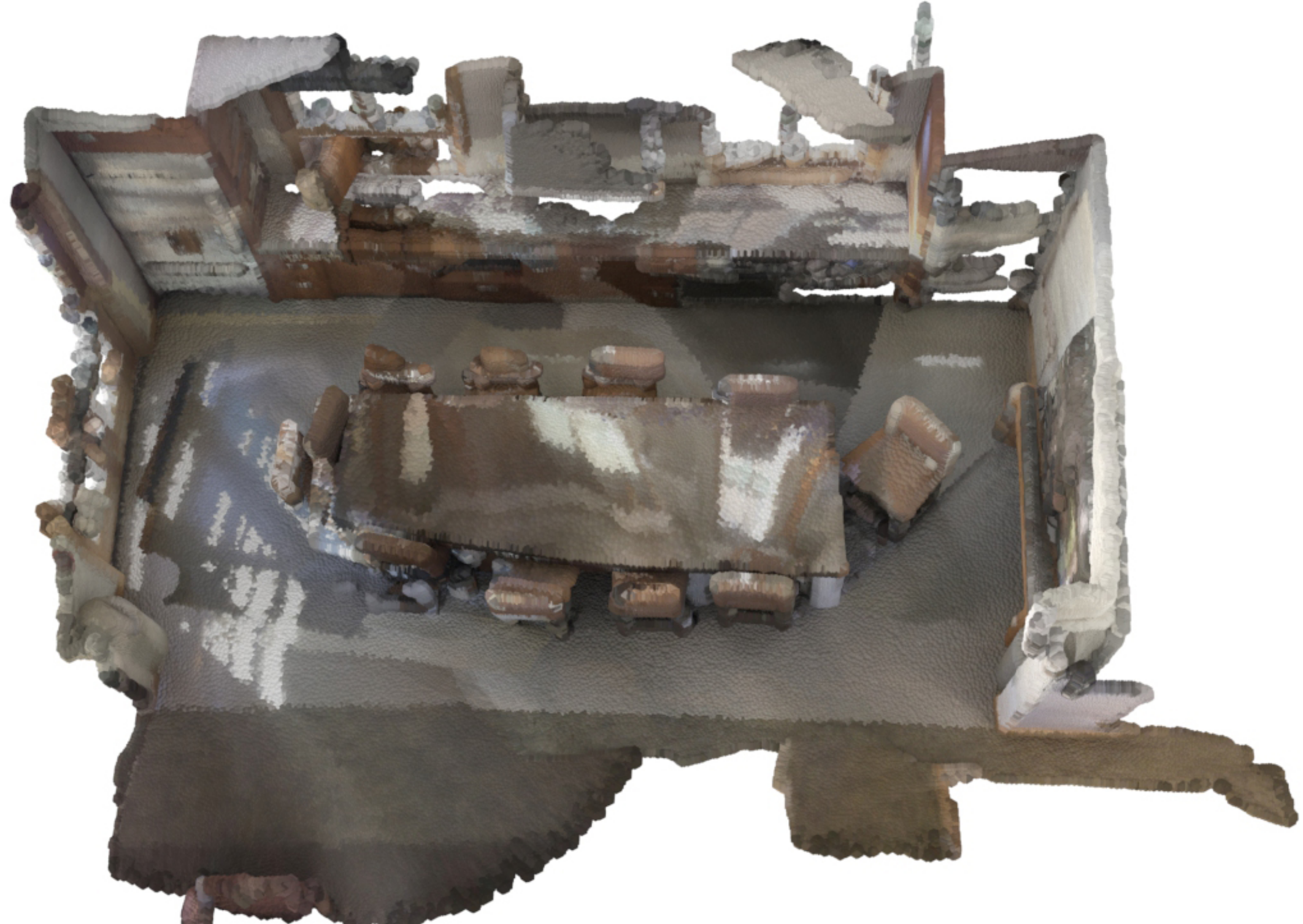}&
            \includegraphics[width=0.3\textwidth]{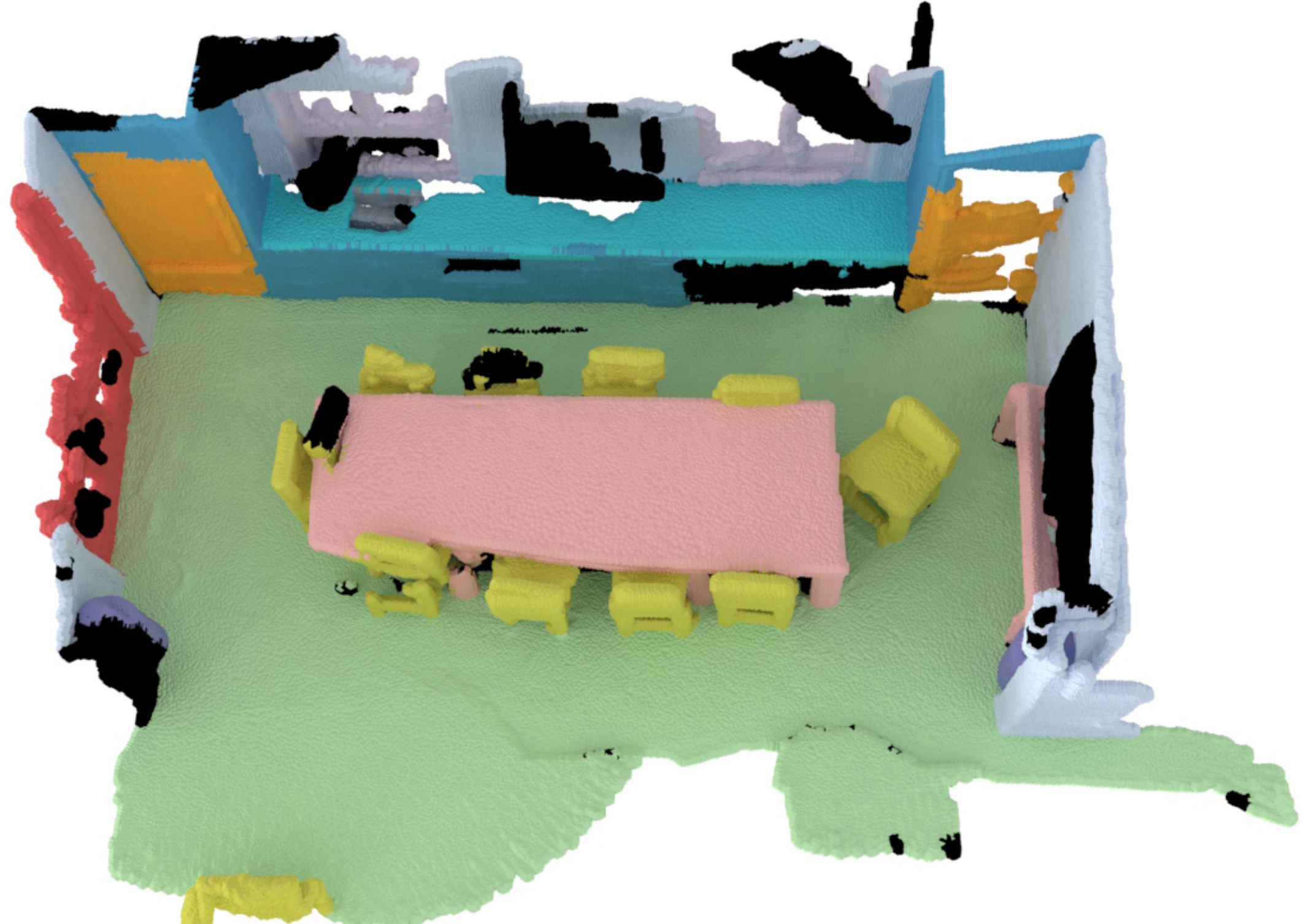}&
            \includegraphics[width=0.3\textwidth]{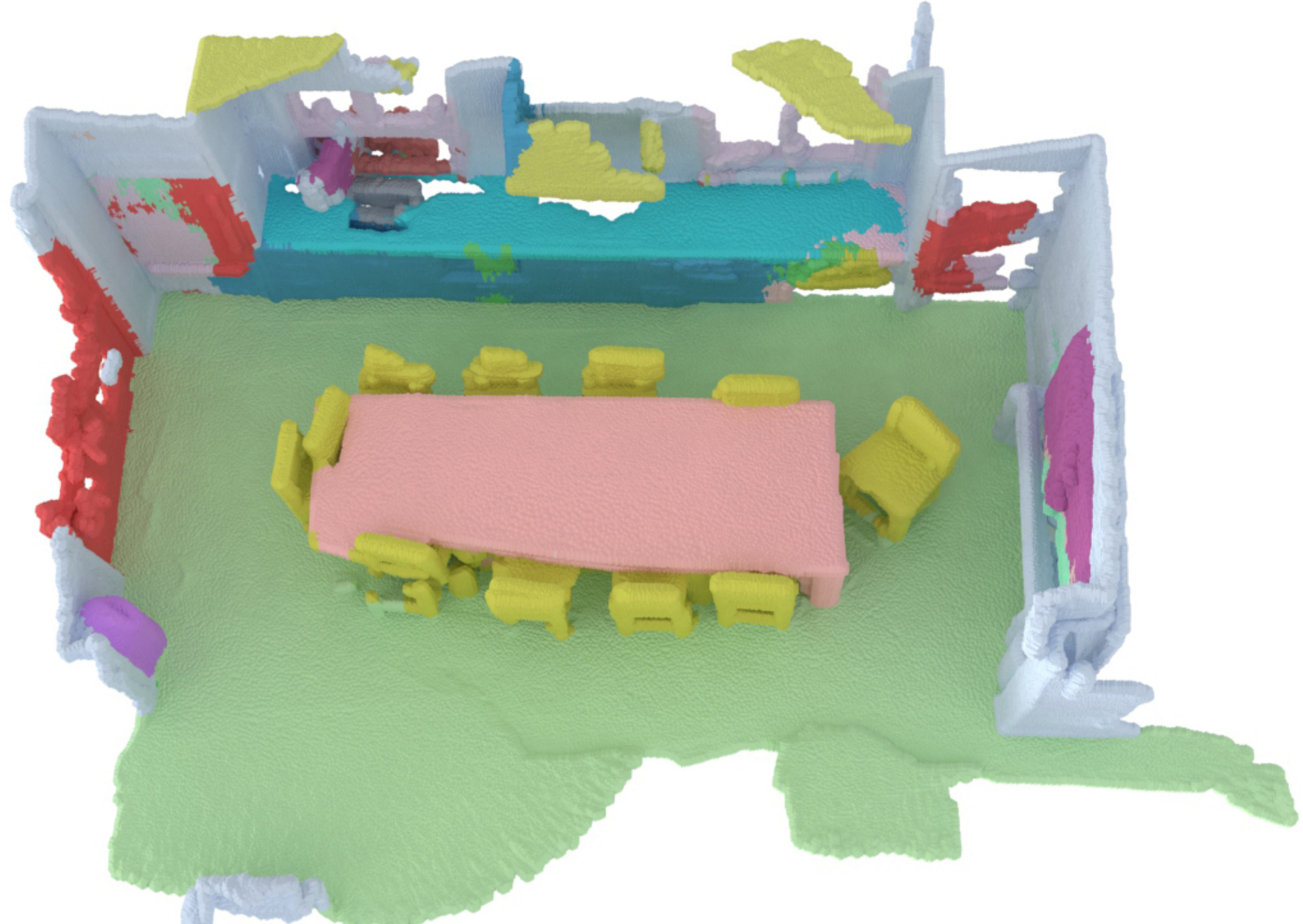}\\
            \multicolumn{5}{c}{
            \includegraphics[width=0.9\textwidth]{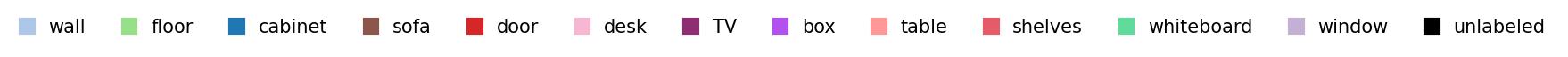}}\\
            \rotatebox{90}{\scriptsize\hspace{20pt}  (b)} & 
            \includegraphics[width=0.3\textwidth]{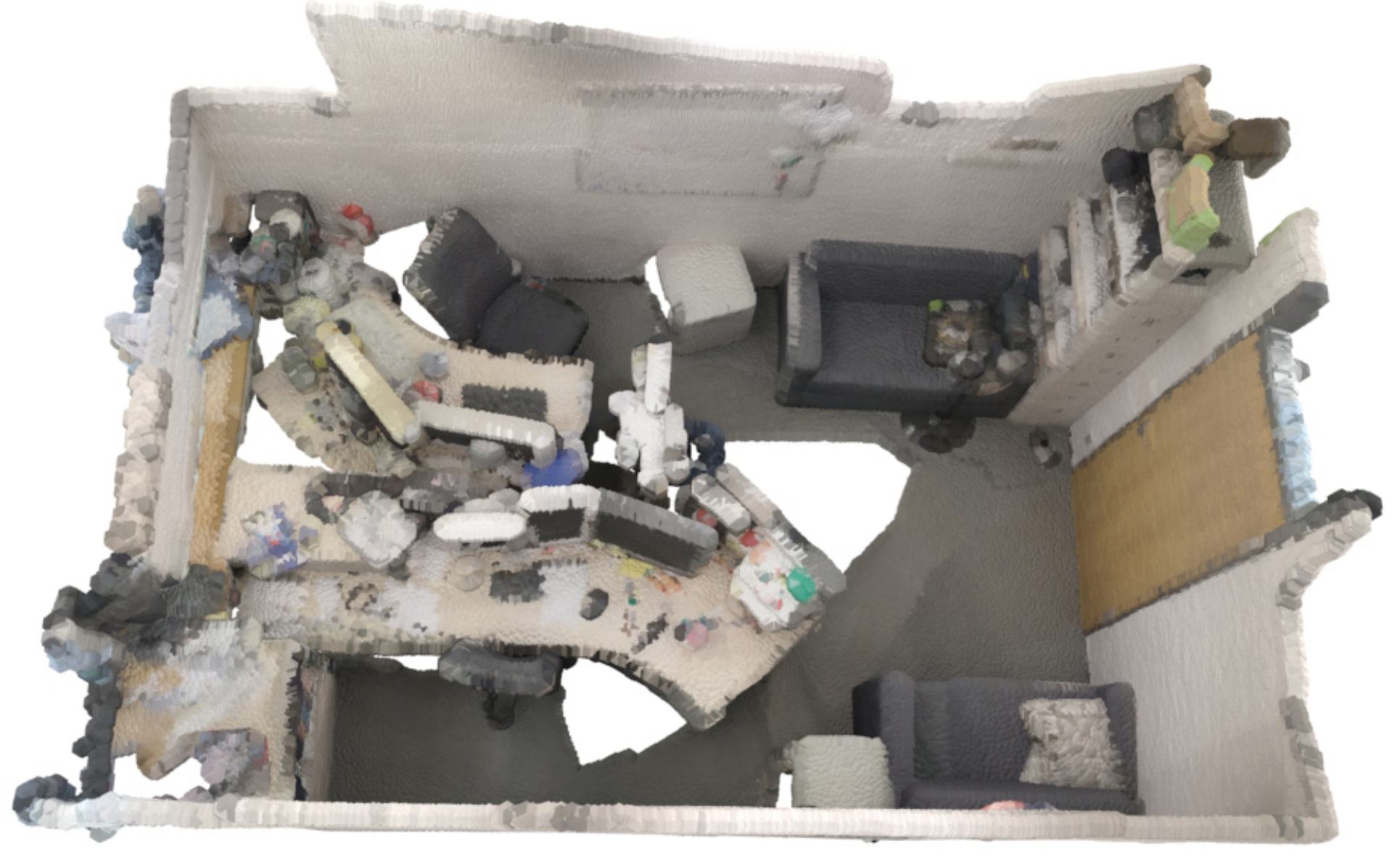}&
            \includegraphics[width=0.3\textwidth]{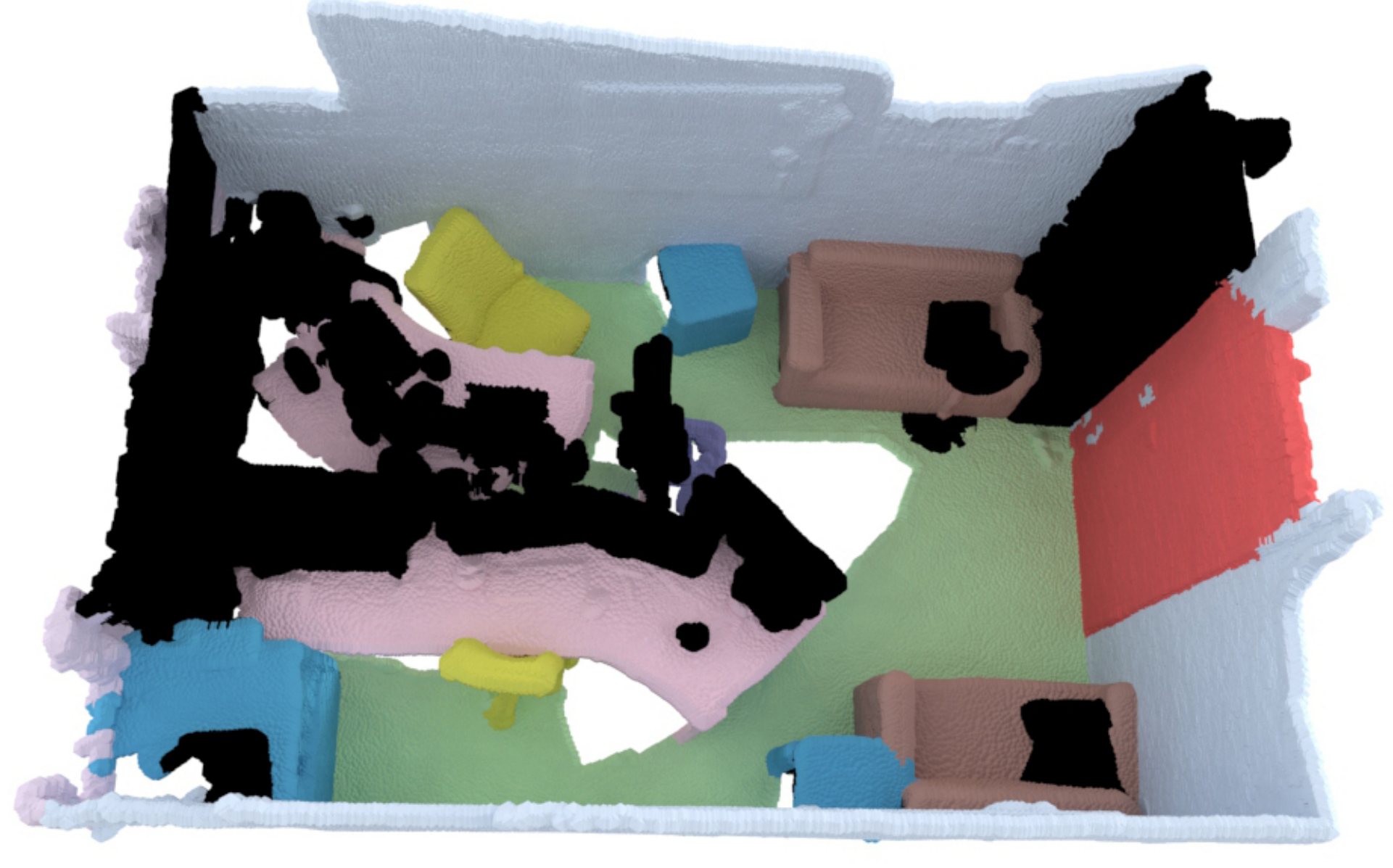}&
            \includegraphics[width=0.3\textwidth]{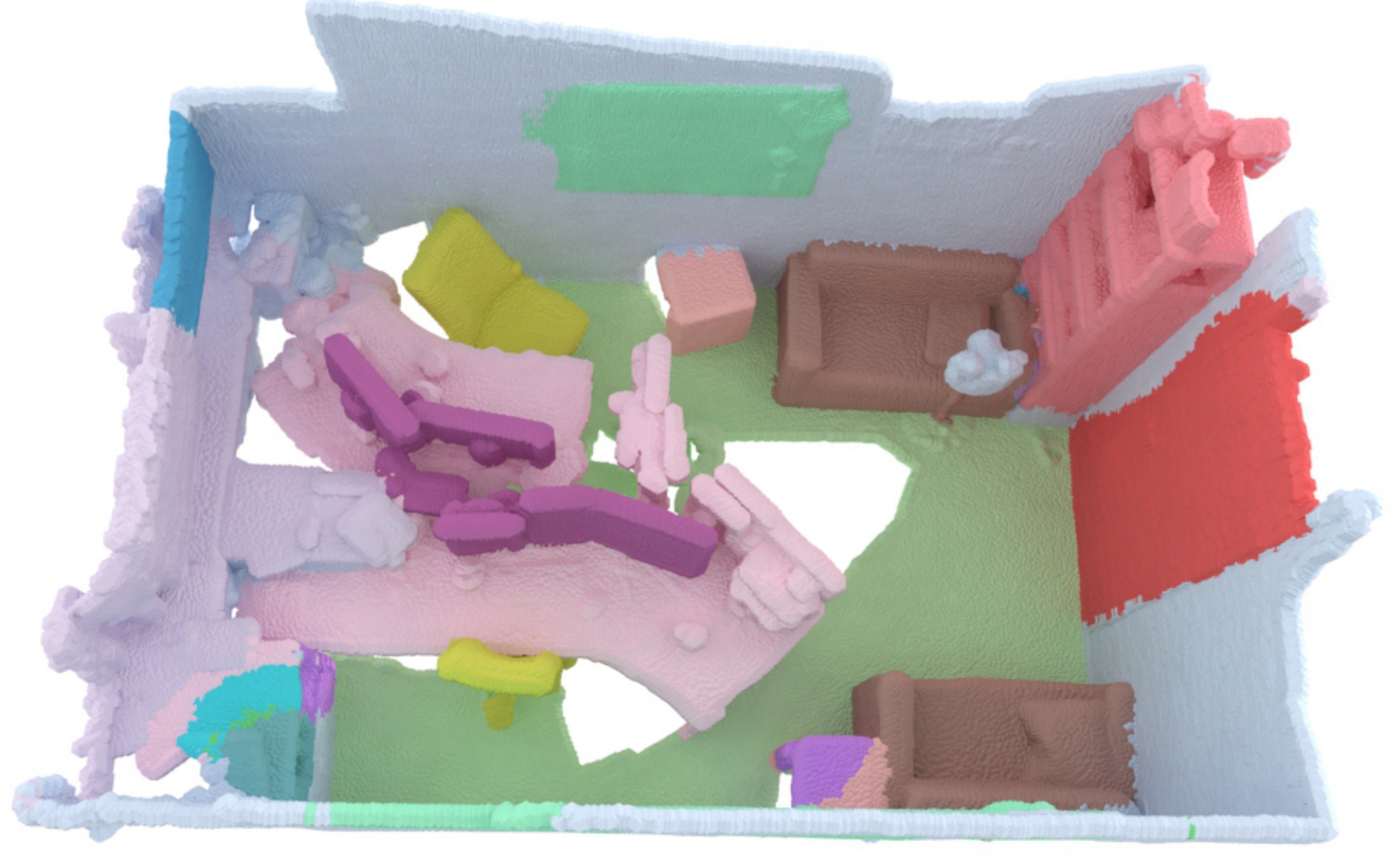}\\
             \multicolumn{5}{c}{
            \includegraphics[width=0.9\textwidth]{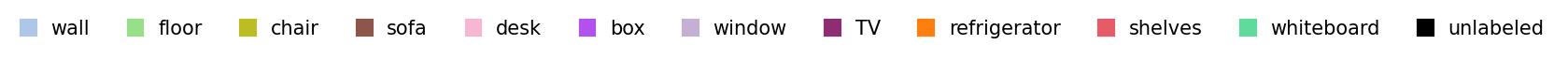}}\\
            \rotatebox{90}{\scriptsize\hspace{20pt}  (c)} & 
            \includegraphics[width=0.3\textwidth]{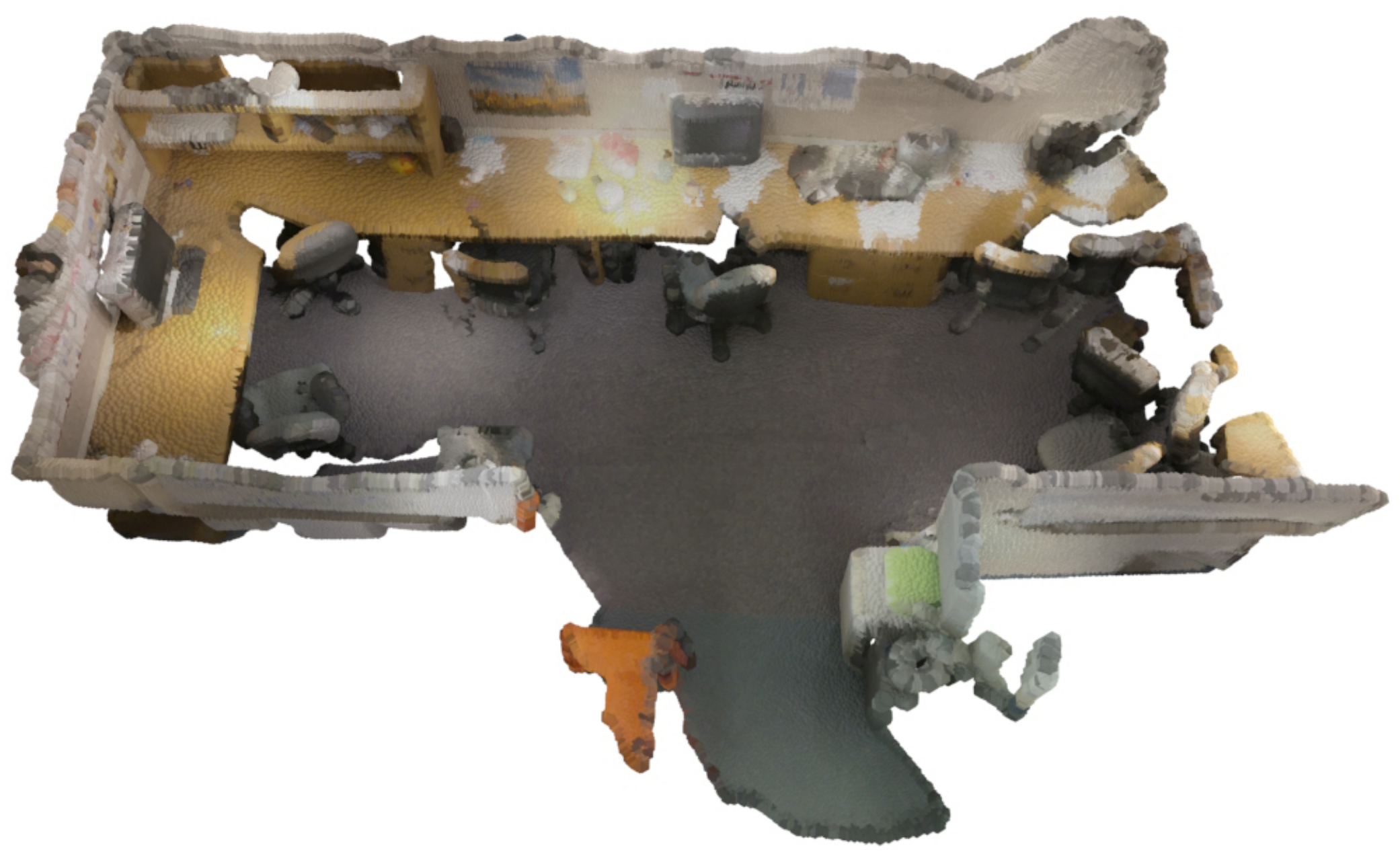}&
            \includegraphics[width=0.3\textwidth]{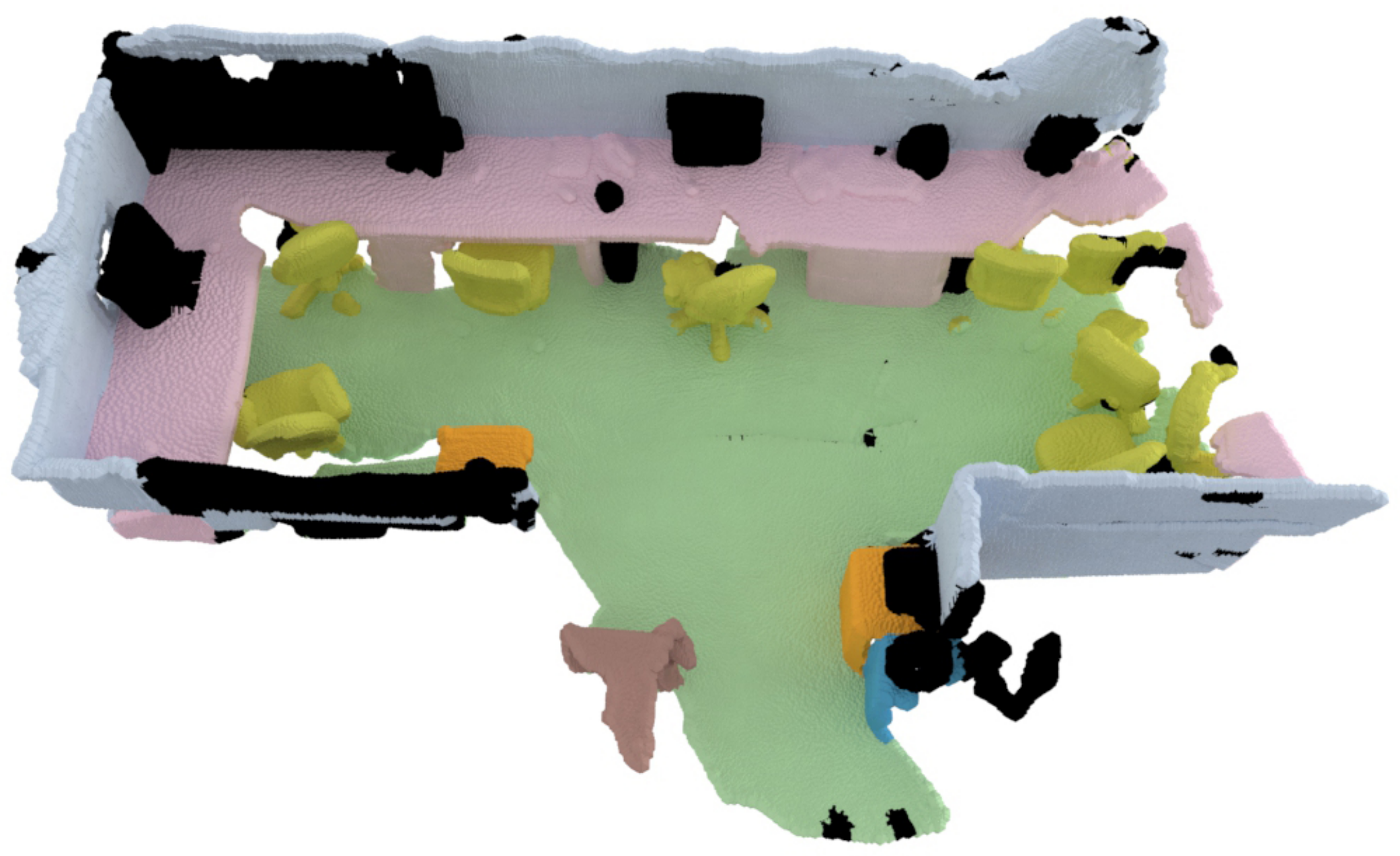}&
            \includegraphics[width=0.3\textwidth]{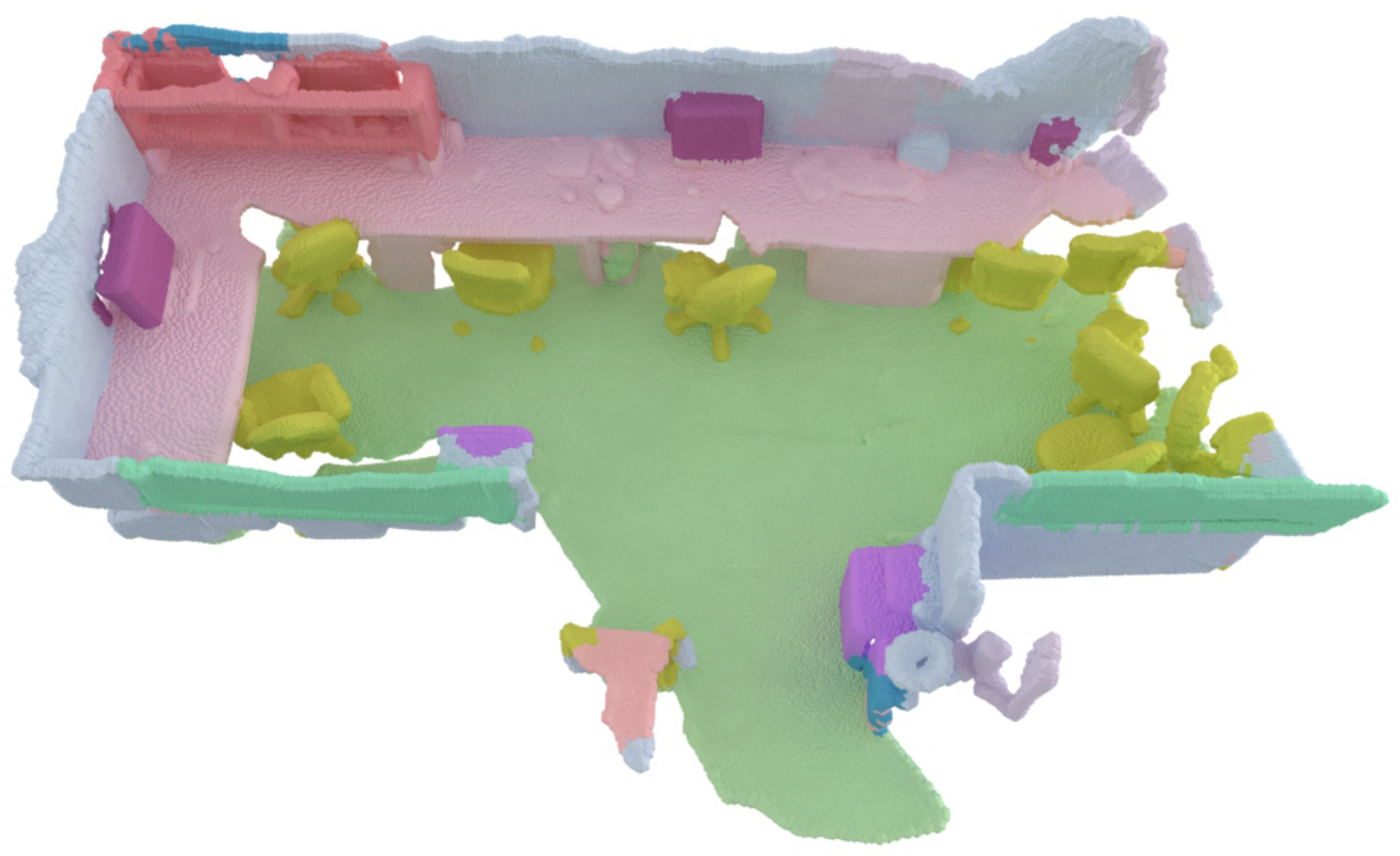}\\
            \multicolumn{5}{c}{
            \includegraphics[width=0.9\textwidth]{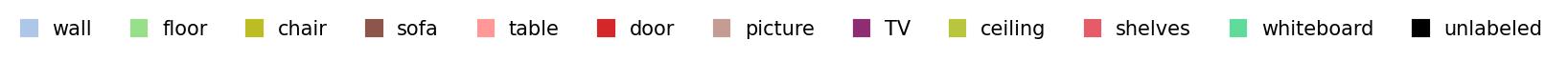}}\\
            \rotatebox{90}{\scriptsize\hspace{20pt}  (d)} & 
            \includegraphics[width=0.3\textwidth]{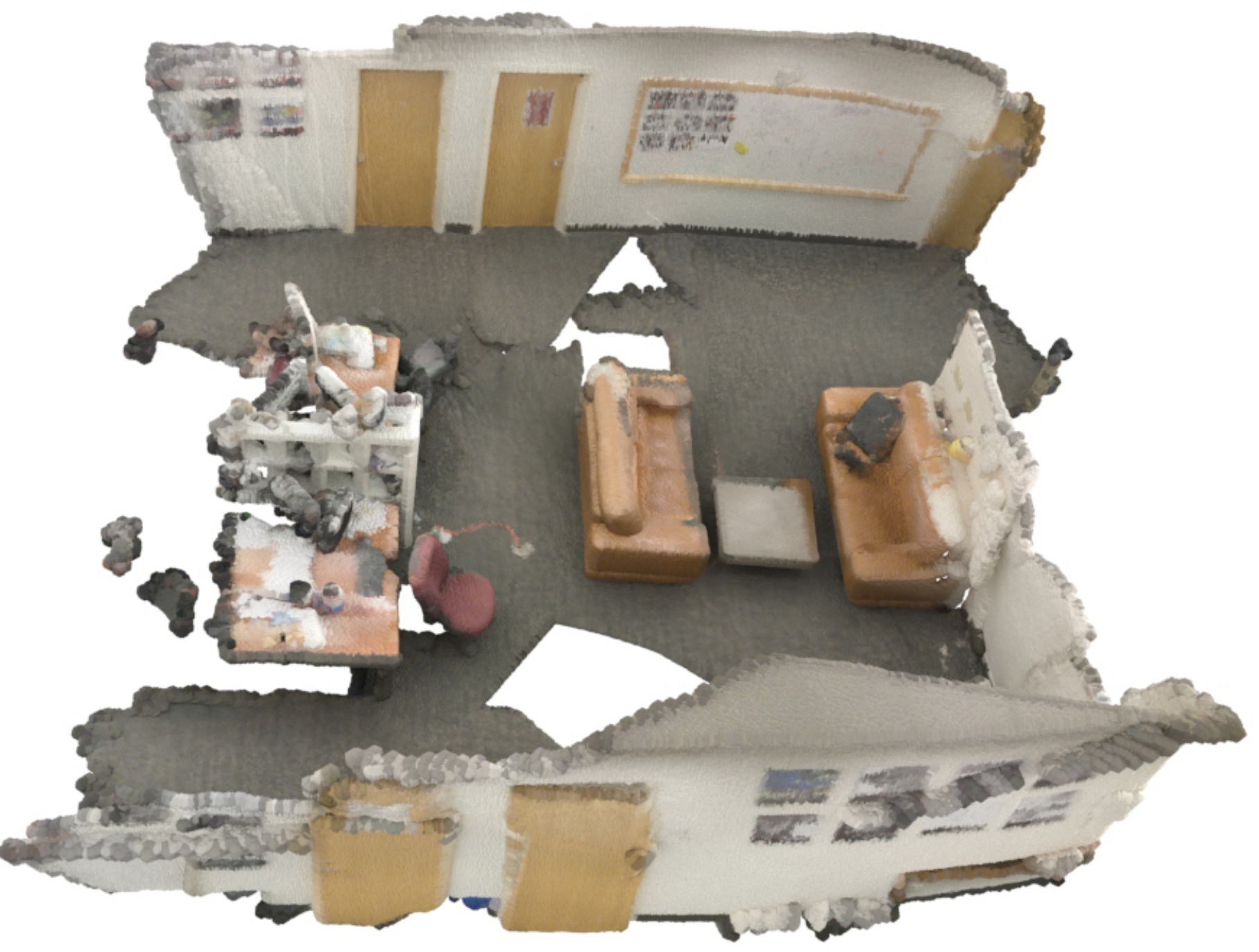}&
            \includegraphics[width=0.3\textwidth]{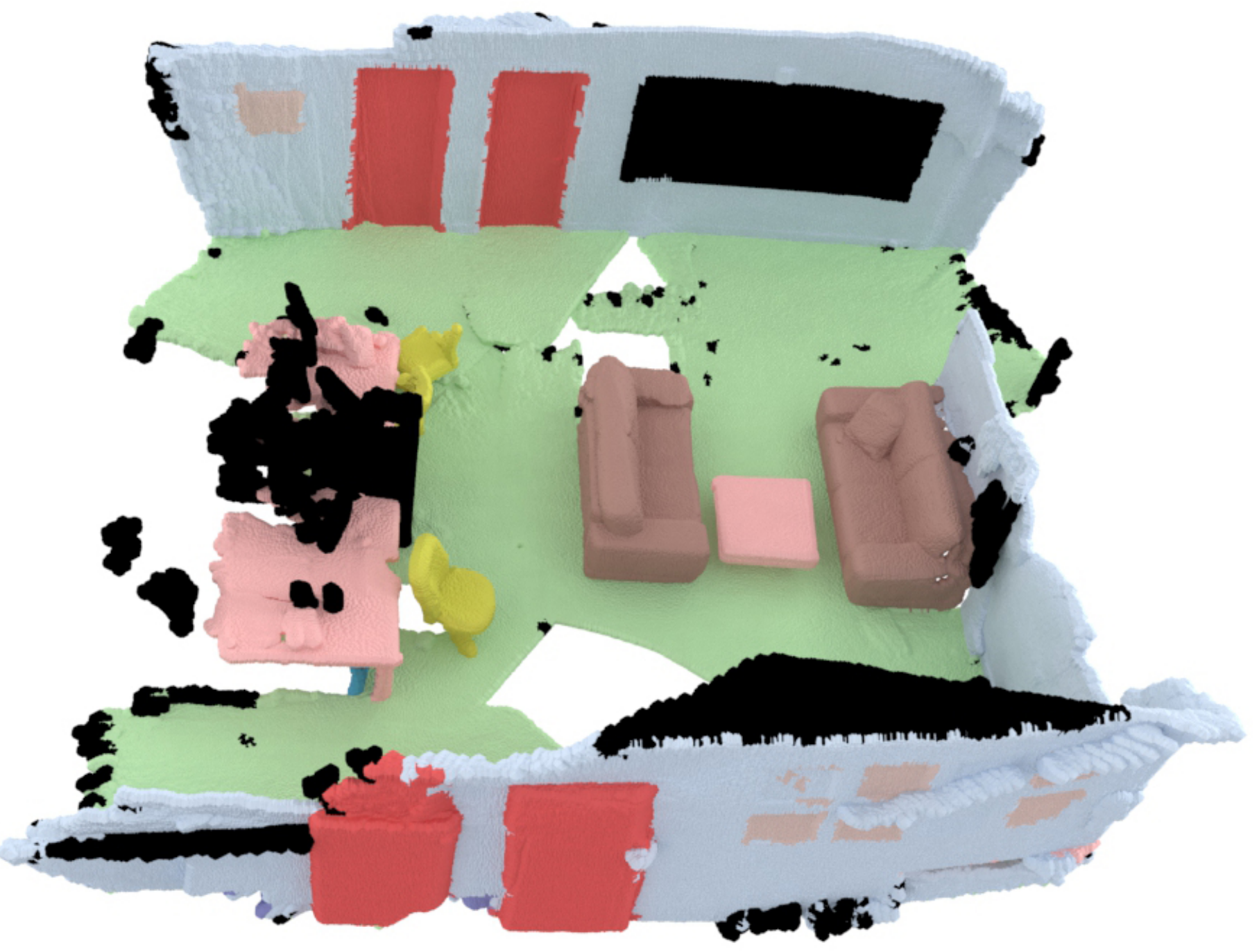}&
            \includegraphics[width=0.3\textwidth]{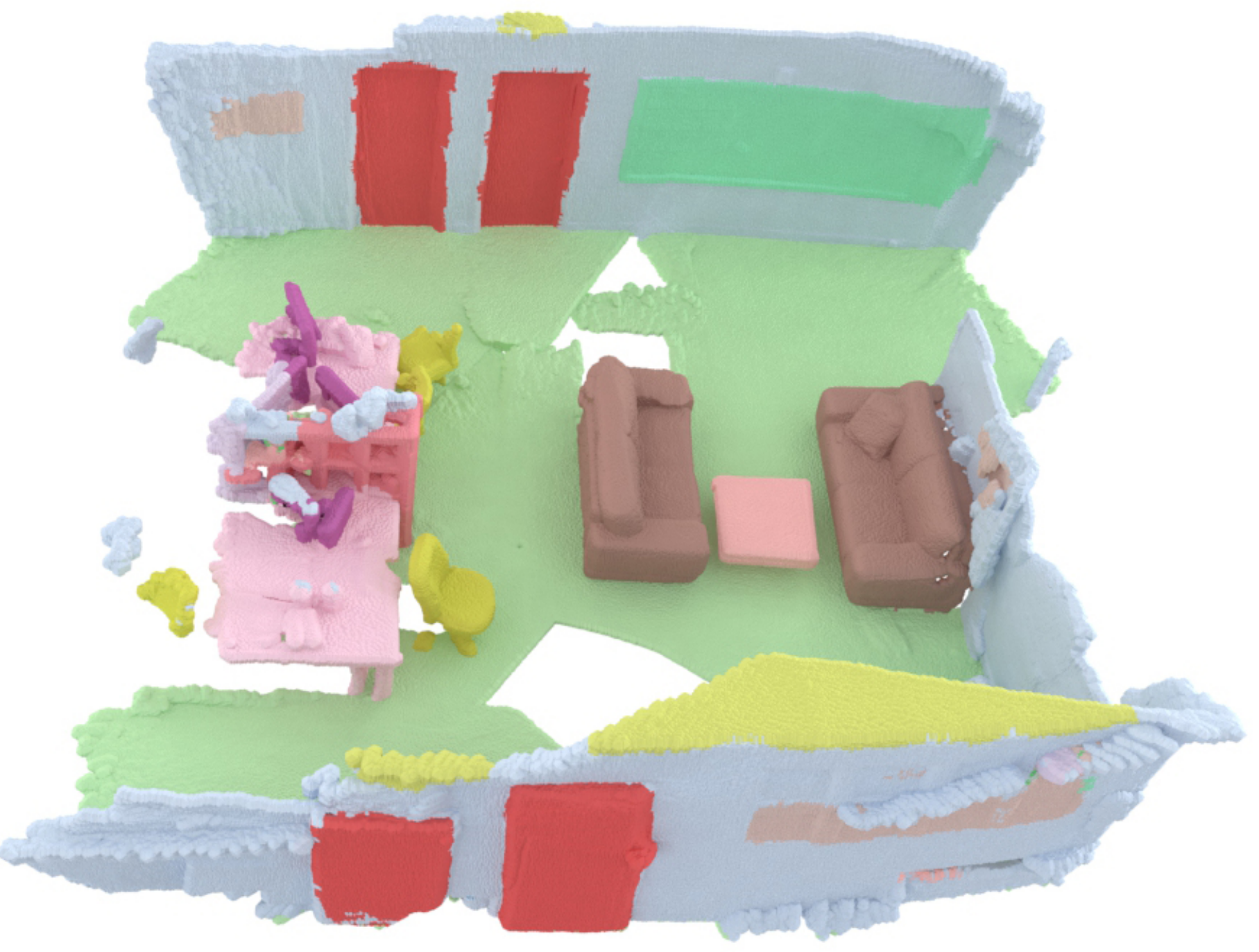}\\
            \\
              {} &{\small \qquad \qquad Input} &
              {\small GT Segmentation}  &
              {\small {Ours}}\\
        \end{tabular}
        \caption{
        \textbf{More qualitative results.} We present more qualitative results of 3D semantic segmentation on ScanNet benchmarks.
        }
        \label{fig:vis}
\end{figure*}

\textbf{Dataset Partitioning for ScanNet with NYU-40 Label.}
To evaluate the open vocabulary capability of GGSD, we expand the original vocabulary size by using the NYU-40 label set. We remove the NYU-40 labels that do not have specific semantics (\eg, ``other structure'', ``other furniture'', ``other prop'') and evenly divide all the rest categories into \textit{Head}, \textit{Common} and \textit{Tail}. 
\textit{Head} classes contain wall, floor, cabinet, bed, chair, bathtub, table, door, toilet, bookshelf, curtain, and ceiling. \textit{Common} classes contain sofa, counter, desk, dresser, refrigerator, shelves, shower curtain, night stand, window, picture, sink and floor mat. \textit{Tail} classes contain blinds, mirror, clothes, pillow, book, box, whiteboard, lamp, towel, bag, person, and television.

\section{More Qualitative Results} \label{sec:B}

We present additional qualitative results in Fig. \ref{fig:vis}. Our method successfully segments all categories that exist in the closed-set ScanNet benchmark, such as {wall}, desk, chair, and {table}. Additionally, our method successfully segments some categories that are not annotated in traditional benchmarks, as shown in Fig. \ref{fig:vis} (a) with examples of ceiling and television, and in Fig. \ref{fig:vis} (b), (c) and (d) with examples of whiteboard and shelves. These qualitative results obtained from GGSD demonstrate its excellent open vocabulary capability.

However, there are also some failure cases in the qualitative results. For example, in Fig. \ref{fig:vis} (b), (c) and (d), computer screens are misclassified as televisions, and in Fig. \ref{fig:vis} (c), the object picture is not segmented. These errors indicate that our model still needs further improvement in fine-grained segmentation and in segmenting objects with less prominent geometric structures, such as pictures. We will explore these areas in future work.

%

{\bibliographystyle{splncs04}
\bibliography{main}